\pdfoutput=1
\documentclass[10pt, logo, twocolumnfalse, copyright, nonumbering]{nvidiatechreport}

\usepackage{booktabs}
\PassOptionsToPackage{numbers,sort&compress}{natbib}

\usepackage{url}
\usepackage{colortbl}
\usepackage{nicefrac}       %
\usepackage{microtype}      %
\usepackage[dvipsnames]{xcolor}         %
\usepackage{adjustbox}
\usepackage{xcolor}
\usepackage{subcaption}    %
\usepackage{float}
\usepackage{stfloats}
\usepackage{listings}
\usepackage{mdframed}
\usepackage{arydshln}
\usepackage{color,soul}
\usepackage{makecell}

\usepackage{amsmath,amssymb,stmaryrd,mathtools}
\usepackage{xspace}
\usepackage{bbm}
\usepackage{mathtools}
\usepackage{bm}
\usepackage{graphicx}
\usepackage{lipsum}
\usepackage{wrapfig}
\usepackage{caption}
\usepackage{multicol, multirow}
\usepackage[table,dvipsnames]{xcolor}
\usepackage{siunitx}
\usepackage{enumitem}
\usepackage{authblk}
\usepackage{hyperref}
\usepackage{cleveref}
\usepackage{booktabs}
\usepackage[most]{tcolorbox}
\usepackage{tabularx}
\usepackage[table]{xcolor}
\usepackage{array}

\usepackage[numbers]{natbib}
\usepackage{chapterbib}

\tcbset{
  colback=gray!5,    % background color
  colframe=black,    % border color
  fonttitle=\bfseries,
  boxrule=0.5pt,
  arc=2mm,
  left=1mm,
  right=1mm,
  top=1mm,
  bottom=1mm,
}
\setlipsum{%
  par-before = \begingroup\color{gray},
  par-after = \endgroup
}

\definecolor{orange(sae/ece)}{rgb}{1.0, 0.49, 0.0}
\definecolor{teal(sae/ece)}{rgb}{0, 0.47, 0.52}
\definecolor{purple}{rgb}{0.74, 0.65, 1.0}
\definecolor{light_gray}{rgb}{0.9, 0.9, 0.9}
\definecolor{medium_gray}{rgb}{0.6, 0.6, 0.6} 
\definecolor{dark_gray}{rgb}{0.2, 0.2, 0.2} 
\definecolor{dark_blue}{rgb}{0.098, 0.239, 0.52}
\definecolor{dark_brown}{rgb}{0.3255, 0.004, 0.001}
\definecolor{dark_purple}{rgb}{0.478, 0.1569, 0.4863}
\definecolor{light_blue}{rgb}{0.33, 0.80, 1}
\definecolor{pastelblue}{RGB}{173,216,230}
\definecolor{pastelyellow}{RGB}{255,253,208}
\definecolor{pastelpink}{RGB}{255,209,220}
\definecolor{pastelgreen}{RGB}{176,226,172}
\definecolor{pastellavender}{RGB}{230,230,250}

\definecolor{NvidiaGreen}{RGB}{118, 185, 0}
\sethlcolor{red!15}

\newcommand{\capalgname}{\textbf{C}ontact Wrench Guidance from \textbf{H}uman Dem\textbf{o}nstration in \textbf{R}obotic \textbf{D}exterous Manipulation\xspace}
\newcommand{\algname}{Contact Wrench Guidance from Human Demonstration in Robotic Dexterous Manipulation\xspace}
\newcommand{\algabrvname}{CHORD\xspace}
\newcommand{\website}{\href{https://nvidia-isaac.github.io/video_to_data/chord/}{\textcolor{magenta}{https://nvidia-isaac.github.io/video\_to\_data/chord/}}}

\newcommand{\numtask}{{4,739}\xspace}
\newcommand{\numtrainedtask}{1,831\xspace}
\newcommand{\successrate}{82.12\%\xspace}

\usepackage{expl3}
\ExplSyntaxOn
\newcommand\latinabbrev[1]{
  \peek_meaning:NTF . {% Same as \@ifnextchar
    #1\@}%
  { \peek_catcode:NTF a {% Check whether next char has same catcode as \'a, i.e., is a letter
      #1.\@ }%
    {#1.\@}}}
\ExplSyntaxOff

\newcommand{\refe}{\mathrm{ref}}
\newcommand{\human}{\mathrm{human}}
\newcommand{\object}{\mathrm{object}}
\newcommand{\robot}{\mathrm{robot}}

\newcommand{\hwrench}{\mathcal{W}_{\mathrm{h},k}}

\newcommand{\hsupport}{\sigma_{\mathrm{h},k}}
\newcommand{\rsupport}{\sigma_{\mathrm{r},k}}

\usepackage{amsmath}

\tcbuselibrary{listings,breakable}

\title{
\makebox[\textwidth][c]{
\parbox{0.95\textwidth}{
\centering
Learning Dexterous Manipulation Using Contact Wrench Guidance From Human Demonstration
}
}
}

\author{
\parbox{\textwidth}{
\centering
\footnotesize
\mbox{Xinghao Zhu$^{*,\ddagger}$},
\mbox{Zixi Liu$^{*}$},
\mbox{Shalin Jain$^{*}$},
\mbox{Chenran Li$^{\dagger}$},
\mbox{Milad Noori$^{\dagger}$},
\mbox{Michael Andres Lin$^{\dagger}$},
\mbox{Huihua Zhao},
\mbox{John Welsh},
\mbox{Mrinal Verghese},
\mbox{Wei Liu},
\mbox{Tingwu Wang},
\mbox{Xingye Da},
\mbox{Zhengyi Luo},
\mbox{Vishal Kulkarni},
\mbox{Naema Bhatti},
\mbox{Yuke Zhu},
\mbox{Linxi Fan},
\mbox{Bowen Wen},
\mbox{Danfei Xu},
\mbox{Soha Pouya},
\mbox{Yan Chang$^{\ddagger}$}

{\normalfont\scriptsize NVIDIA}

{\normalfont\scriptsize
$^{*}$Equal Contribution \quad
$^{\dagger}$Core Contributor \quad
$^{\ddagger}$Project Lead and Corresponding Author
}
}
}

\begin{abstract}
\textbf{Abstract:}
Dexterous robot manipulation can benefit from the abundance of human demonstrations, but transferring such demonstrations to robot policies remains challenging. We present \algname{} (\algabrvname{}), a framework for long-horizon manipulation of rigid and articulated objects with reinforcement learning. The key idea is object-centric contact wrench space guidance: we represent human and robot motions by the forces and torques they can induce on the object, enabling similarity to be measured by the induced instantaneous motions. This guidance makes reinforcement learning more scalable for contact-rich dexterous manipulation. We further introduce a large-scale simulation benchmark with \numtask bimanual dexterous manipulation tasks, constructed from motion-capture datasets and reconstructed in-house videos. Evaluated on \numtrainedtask benchmark tasks, \algabrvname{} achieves an average success rate of \successrate, demonstrating strong scalability. \algabrvname{} also generalizes to whole-body manipulation from hand-only and third-person demonstrations, achieving a 90.77\% success rate, and the learned policies transfer to the real world in both open-loop and closed-loop settings. Videos and code are available at \website.
\end{abstract}

\begin{document}

\maketitle

\section{Introduction}
\label{sec:intro}

Robotic dexterous hands offer a compelling platform for general-purpose manipulation because of their morphological similarity to human hands, enabling robots to exhibit human-level dexterity.
Leveraging human demonstrations has shown promise in advancing dexterous manipulation by mitigating the exploration challenges of policy optimization~\cite{mandi2026dexmachina,li2025maniptrans,pan2025spider,liu2025quasisim,liu2025dextrack,dexh2r} and supporting the learning of transferable human-robot representations~\cite{egomimic,punamiya2026egobridge,egoscale,wen2022you,hsu2025spot}.
However, existing methods that use human demonstrations for optimization often rely on brittle assumptions for how the demonstration should be transferred, while representation-learning approaches typically require aligned human-robot data for each task and object, limiting their scalability beyond curated settings.
This motivates %the central question of 
our work: \emph{How can a robot learn transferable and scalable task knowledge from human demonstrations?}

A dexterous robot cannot simply replay human hand motion to reproduce human manipulation behavior~\cite{mandi2026dexmachina,dexh2r,egoscale}. Differences in morphology, kinematics, and hand geometries require the robot to use different motions and contacts to achieve the same functional effect on the object. 
% This issue is especially pronounced in long-horizon, contact-rich manipulation, where success depends on a sequence of distinct contact modes and small errors in contact timing or mechanics can lead to large deviations in object motion.
Our key insight is that contact provides a natural bridge between human demonstrations and robot actions. Rather than matching 3D human contact locations~\cite{mandi2026dexmachina,li2025maniptrans,pan2025spider}, the robot should find contacts that induce object motions aligned with the human demonstration. 
% To this end, we represent contacts in an object-centric wrench space, which captures the force-torque directions and instantaneous object motions a contact configuration can physically support~\cite{murray2017mathematical}. 
To this end, we represent contacts in an object-centric wrench space, capturing the force-torque directions and object motions supported by each contact configuration~\cite{murray2017mathematical}.
This representation is shared across embodiments: humans and robots may differ in contact locations, contact counts, and hand morphology, but their contacts can still be compared in wrench space with respect to the object-level motions they are capable of inducing.

Concretely, we present \capalgname (\algabrvname), a framework for learning dexterous, contact-rich manipulation from human demonstrations. Given a human reference, \algabrvname casts cross-embodiment imitation as reproducing the demonstrated object-state evolution through physically consistent interactions expressed in contact-wrench space, rather than directly matching embodiment specific human motion. This formulation enables long-horizon RL across diverse rigid and articulated manipulation tasks without task-specific robot demonstrations.
% We also introduce a large-scale simulation benchmark of \numtask bimanual dexterous manipulation tasks, covering rigid-, articulated-, and multi-object interactions from open-source motion-capture datasets~\cite{fan2023arctic,liu2024taco,hot3d,oakink} and reconstructed in-house videos. Across \numtrainedtask benchmark tasks, \algabrvname achieves an average success rate of \successrate and outperforms prior baselines. The same contact-guided formulation further extends to whole-body manipulation from hand-only and third-person demonstrations, reaching 90.77\% success rate, and transfers to a real bimanual dexterous robot in open-loop and closed-loop settings.
We further introduce a large-scale simulation benchmark of \numtask bimanual dexterous manipulation tasks, built from open-source motion-capture datasets~\cite{fan2023arctic,liu2024taco,hot3d,oakink} and reconstructed in-house videos. Across \numtrainedtask tasks, \algabrvname achieves an average success rate of \successrate, outperforming prior baselines. The same formulation extends to whole-body manipulation from hand-only and third-person demonstrations, and transfers to a real bimanual dexterous robot in both open-loop and closed-loop settings.

% Our contributions are summarized as follows:
% \begin{itemize}[leftmargin=*, itemsep=1pt, topsep=-2pt, parsep=0pt]
% \item We propose \algabrvname, a wrench-space contact-guided framework for transferring human demonstrations to dexterous robot policies with reinforcement learning.
% \item We introduce a benchmark of \numtask long-horizon bimanual dexterous manipulation tasks with human demonstrations spanning rigid, articulated, and multi-object interactions.
% \item We validate \algabrvname in large-scale simulation, showing \successrate success rate on \numtrainedtask tasks and the generalization to whole-body with a 90.77\% success rate. We also show that the learned policies can be successfully deployed on real hardware.
% \end{itemize}

Our contributions are: \textbf{(1)} \algabrvname, a wrench-space contact-guided framework for transferring human demonstrations to dexterous robot policies with reinforcement learning; \textbf{(2)} a benchmark of \numtask long-horizon bimanual dexterous manipulation tasks spanning rigid, articulated, and multi-object interactions; and \textbf{(3)} validation of \algabrvname in large-scale simulation, achieving a \successrate success rate on \numtrainedtask tasks and 90.77\% success in whole-body manipulation tasks with $\sim$2 hours of training on a single GPU, with successful deployment on real hardware.

\section{Related Work}
\label{sec:related_work}

% 1. Human demonstration is useful for robot learning. For humanoid whole body control and transferable human-to-robot representations training. Dexterous manipulation has also been used in human demonstrations for learning. But dexterous manipulation is hard because of the large action space and non-smoothness of the optimization contour.
Learning from human demonstration is an increasingly prominent approach for enabling robots to acquire skills from existing experts~\cite{ravichandar2020recent}, demonstrating success in learning whole-body control~\cite{ze2025twist, liao2025beyondmimicmotiontrackingversatile, luo2025sonic} and representation learning~\cite{nair2022r3m, egoscale, egomimic, punamiya2026egobridge}. However, dexterous manipulation with human demonstration remains difficult due to high-dimensional hand actions and non-smooth contact dynamics, where small motion changes can abruptly change object behavior~\cite{okamura2000overview, antonova2022rethinking,pang2023global,zhu2023difflfd,suh2026contacttrustregion,suh2022bundle}.
% 2. How did previous works solve dexterous manipulation with human demonstration?
Prior work uses reinforcement learning in simulation~\cite{qin2021dexmv, chen2024object, han2024learning, pmlr-v305-lum25a}. These methods rely on task-level imitation objectives such as object and hand-reference tracking, while recent approaches add contact guidance to improve exploration. ManipTrans~\cite{li2025maniptrans} rewards contact forces near demonstrated hand-object interactions, whereas DexMachina~\cite{mandi2026dexmachina} and SPIDER~\cite{pan2025spider} use demonstrated \emph{contact locations} together with curricula such as virtual object control. The virtual object controller (VOC) applies an auxiliary wrench to move the object along the reference trajectory during early training. This makes the manipulation problem smoother by reducing the need for the policy to immediately discover precise contact timing and contact forces. Instead of receiving a useful object-tracking reward only after successful contact, the policy observes a denser learning signal.

\begin{wrapfigure}{r}{0.25\textwidth}
    \vspace{-3em}
    \centering
    \includegraphics[width=\linewidth]{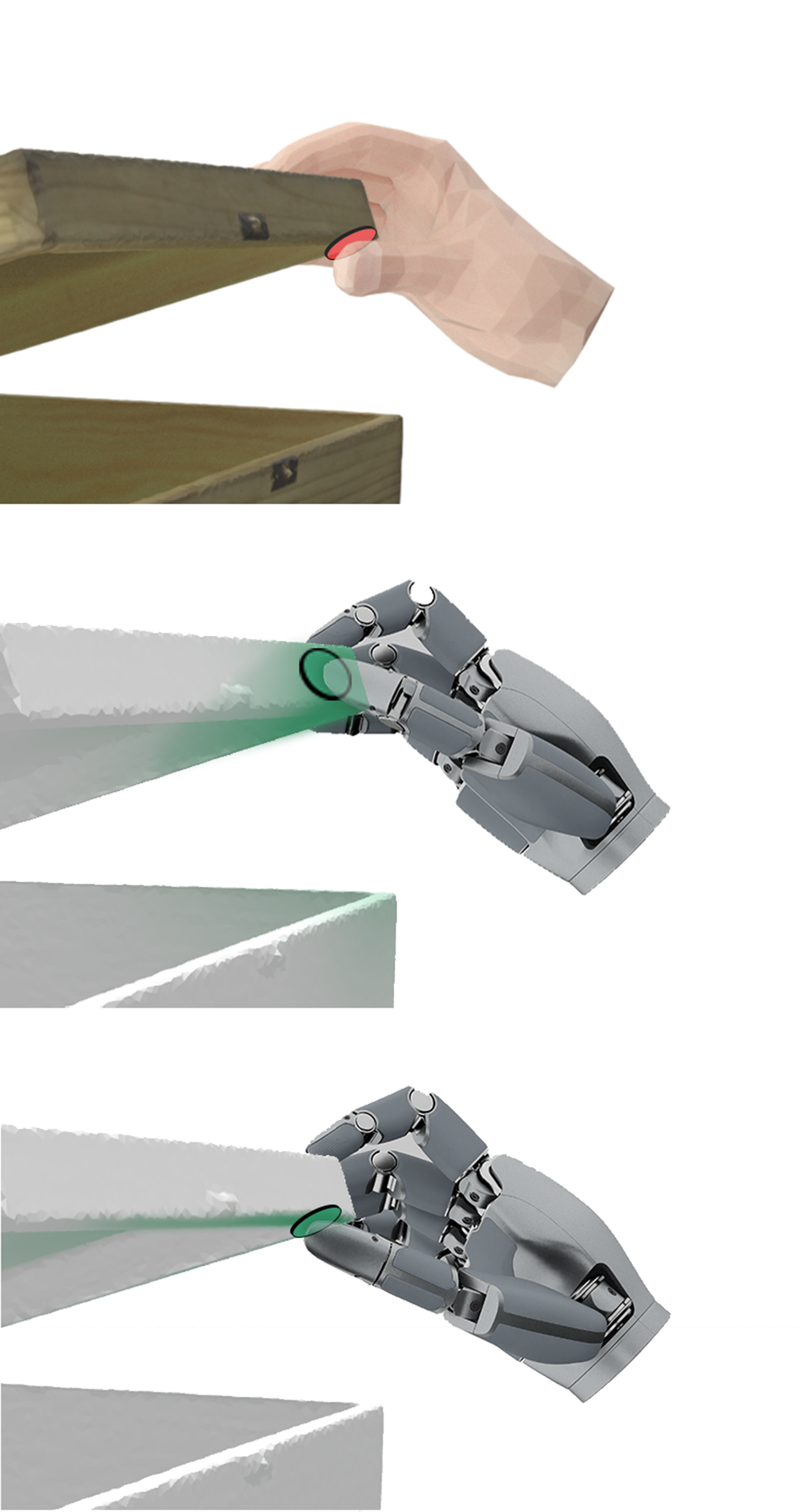}
    \label{fig:cws_reward}
    \vspace{-3em}
\end{wrapfigure}

% 3. What we think about the contact wrench and how previous work uses it.
% While the virtual object controller eases exploration, they introduce local optima to the training. The robot might not learn effective contact behavior with the object controller active.
The VOC eases exploration; however, it can introduce local optima, leading to ineffective contact behavior.
Specifically, the contact location-based reward~\cite{mandi2026dexmachina,pan2025spider} encourages the robot to touch object regions similar to those touched by the human demonstrator. However, contact location alone does not determine the effect of a contact (\Cref{fig:cws_reward}). The same object region can produce different object motions depending on the contact surface normal and force direction. An alternative is to supervise contact in \textit{wrench space}, the space of forces and torques that contacts can apply to the object~\cite{bicchi1995closure,ferrari1992planning,murray2017mathematical,zhu2023difflfd}. 
Contact wrench rewards in prior RL work~\cite{koenig2022role, merzic2019leveraging, melnik2019tactile, melnik2021using}, optimize for static stability, making them effective for grasping but too rigid for general robot manipulation.
In contrast, we leverage the wrench space, for the first time, to measure the similarity between a human demonstration and a robot execution in terms of the \emph{induced motion} from contact in the RL training. 
This allows contacts with different locations, contact counts, or hand morphologies to be compared by their object-level mechanical effects during RL training. Crucially, this metric accommodates transient, non-force-closure phases of long-horizon manipulation such as pushing, levering, and sliding, as well as the manipulation of articulated objects.

\section{Method}

\label{sec:method}

\begin{figure}
    \vspace{-0.5em}
    \centering
    \includegraphics[width=\linewidth]{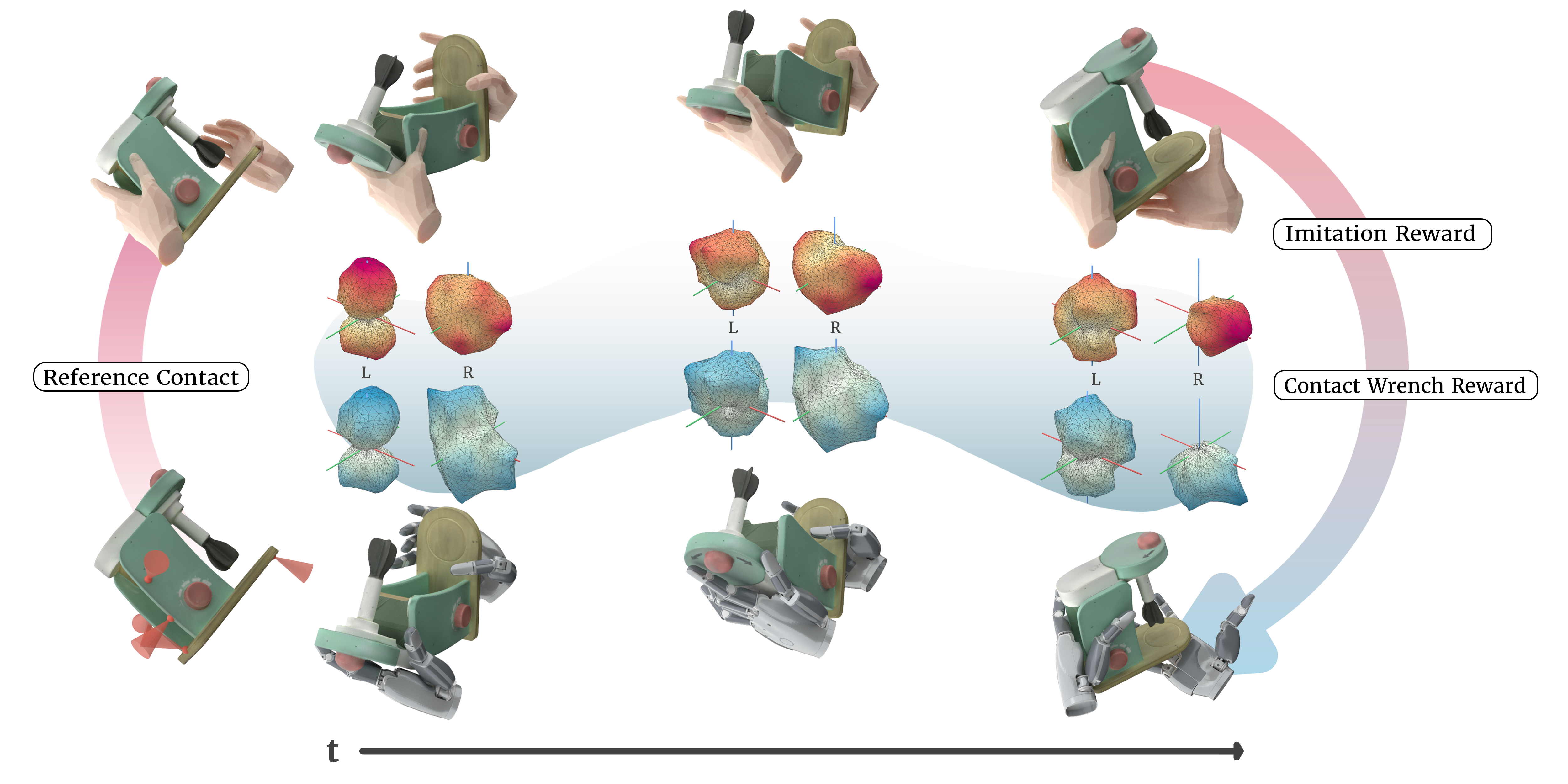}
    \caption{\textbf{\algabrvname{}} combines imitation and contact guidance rewards for RL training. \textit{Left:} Contact wrench references extracted from the human demonstration, with the corresponding contact positions and friction cones (red) in the lower panel. \textit{Top:} Human demonstration of a mixer-closing task. \textit{Middle:} Evolution of per-hand contact wrenches, visualized with force manifolds, throughout the task, where red denotes the human contact wrench, and blue denotes the corresponding contact wrench generated by (\textit{Bottom}) \algabrvname{}-trained policy.
    }
    \label{fig:method-draft}
    \vspace{-0.5em}
\end{figure}

% Learning from Human Overview
Our goal is to learn long-horizon manipulation of rigid and articulated objects from human demonstrations. We consider objects with $K$ rigid bodies, either as $K$ separate rigid objects or as an articulated object with $K$ parts. Each demonstration provides a reference trajectory $\tau^{\refe}=\{x^{\human}_t, x^{\object}_t\}_{t=1}^H$, where $x^{\human}_t$ denotes 3D human hand keypoints and $x^{\object}_t=\{x^{\object,k}_t\}_{k=1}^K$ denotes object-part poses with $x^{\object,k}_t \in \mathrm{SE}(3)$ for each part (\Cref{fig:method-draft}). We begin by retargeting the hand motion by solving inverse kinematics from human keypoints to robot configurations $x^{\mathrm{robot}}_t$. The policy $\pi(a_t \mid o^{\robot}_t, o^{\object}_t; x^{\mathrm{robot}}_t, x^{\object}_t)$ then predicts actions from current robot observations $o^{\robot}_t$, object observations $o^{\object}_t$ and reference motion $x^{\mathrm{robot}}_t, x^{\object}_t$, with the objective that rollout object poses $\{s^{\object}_t\}_{t=1}^H$ track the reference trajectory $\{x^{\object}_t\}_{t=1}^H$.
% under the system dynamics $(s^{\mathrm{robot}}_{t+1}, s^{\object}_{t+1})=f(s^{\mathrm{robot}}_t, s^{\object}_t, a_t)$, 

We present \algname{} (\algabrvname{}), a framework for long-horizon, contact-rich manipulation of rigid, articulated objects, and multi-object interactions. We first formulate the reinforcement learning rewards (\Cref{subsec:contact_reward}). We then describe how \algabrvname{} enables efficient and robust learning from noisy human references, and how it extends to whole-body manipulation (\Cref{subsec:general}). Finally, we introduce our dexterous manipulation benchmark (\Cref{subsec:benchmark}). More details on the MDP setup, our insights into contact guidance, and smooth contact dynamics can be found in the Appendix.

\subsection{RL with Wrench Space Contact Guidance}
\label{subsec:contact_reward}

We train the policy with task-tracking, motion-imitation, and contact-guidance rewards, $r = r_{\mathrm{task}} + r_{\mathrm{imit}} + r_{\mathrm{contact}}$ (Figure~\ref{fig:method-draft}) with the VOC~\cite{mandi2026dexmachina}. 
The task reward encourages the rollout to follow the reference object motion, including both object-part poses and inter-part relations: $r_{\mathrm{task}} = \exp\left(-\frac{\sum_{k=1}^{K}\|x^{\mathrm{object},k}_t \ominus s^{\mathrm{object},k}_t\|_2^2}{\textrm{var}_{\mathrm{obj}}}\right) + r_{\mathrm{relative}}$, where $\ominus$ denotes the pose difference in $\mathrm{SE}(3)$. 
For multi-object interaction, tracking each object part independently is often insufficient for tasks that depend on precise object-object geometry, such as insertion, pouring, scooping, and tool use. We therefore add a relative reward, $r_{\mathrm{relative}}=m(t)\exp\left(-\frac{e_{\object}}{\textrm{var}_{\mathrm{rel}}}\right)$, where $m(t)$ activates the term only during object-object interaction phases, and $e_{\object}$ measures the inter-object pose error. 
The imitation reward regularizes the robot toward the retargeted human motion, $r_{\mathrm{imit}} = \exp\left(-\frac{\|x^{\mathrm{robot}}_t - s^{\mathrm{robot}}_t\|_2^2}{\textrm{var}_{\mathrm{imit}}}\right)$. $\textrm{var}_{\mathrm{obj}}, \textrm{var}_{\mathrm{rel}}, \textrm{var}_{\mathrm{imit}}$ are variances for the exponential kernel. % See Appendix for more details. 

% Induced Wrench, Wrench Matrix, and Support Functions
As discussed in~\Cref{sec:related_work}, location-based contact rewards are often insufficient. In the right-side box-opening task, the top row shows the human demonstration, where the thumb contacts the underside of the lid. The middle-row heatmap visualizes the guidance score assigned to each candidate contact point by a location-based contact reward. Under this reward, the optimized contact is spatially close to the human contact but it is physically mismatched: the contact normal is nearly perpendicular to that of the demonstration. In contrast, the bottom-row heatmap shows the guidance using our proposed contact wrench-space reward, which encourages contacts capable of producing object motion consistent with the human demons tration.

At each time step, for each rigid part $k$, we extract $c_{\mathrm{h},k}$ contact positions $p_{\mathrm{h},k}=\{p^i_{\mathrm{h},k}\}_{i=1}^{c_{\mathrm{h},k}}$ and contact normals $n_{\mathrm{h},k}=\{n^i_{\mathrm{h},k}\}_{i=1}^{c_{\mathrm{h},k}}$ from the human demonstration $\tau^{\mathrm{ref}}$ in the object frame~\cite{mandi2026dexmachina}. For any feasible contact force $f^j_{\mathrm{h},k}\in\mathbb{R}^3$ at contact $i$, the induced primitive wrench is defined as $w^{i,j}_{\mathrm{h},k}=\begin{bmatrix} f^j_{\mathrm{h},k}, & p^i_{\mathrm{h},k}\times f^j_{\mathrm{h},k} \end{bmatrix}^{\top}\in\mathbb{R}^6$.

To characterize the contact wrench geometry induced by $c_{\mathrm{h},k}$ human contact points, we approximate the Coulomb friction cone at each contact with a polyhedral cone containing $d$ unit-magnitude edge forces. The $j$-th edge force at contact $i$ induces a primitive wrench $w^{i,j}_{\mathrm{h},k}$. We collect all primitive wrenches for the hand--part pair into the wrench matrix:
\begin{equation}
\label{eq:wrench_matrix}
\hwrench = \begin{bmatrix}
w^{1,1}_{\mathrm{h},k} & \cdots & w^{1,d}_{\mathrm{h},k} &
\cdots &
w^{c_{\mathrm{h},k},1}_{\mathrm{h},k} & \cdots & w^{c_{\mathrm{h},k},d}_{\mathrm{h},k}
\end{bmatrix}
\in \mathbb{R}^{6\times (c_{\mathrm{h},k} d)} .
\end{equation}

The wrench matrix $\hwrench$ represents the force--torque directions that the human contacts can apply to object part $k$, and therefore captures their motion-inducing affordance. However, directly comparing two wrench matrices, such as those from a human and a robot, is difficult because they may have different numbers of columns and arbitrary primitive-wrench orderings.
We therefore propose to compare their induced wrench geometry using a support function. Given $b$ pre-sampled unit directions $\mathcal{B}\in\mathbb{R}^{6\times b}$, we define the support function of $\hwrench$ as \begin{equation}
\label{eq:support_function}
\hsupport =
\max_{\mathrm{col}}
\left(
\mathcal{B}^\top \hwrench
\right)
\in \mathbb{R}^{b}
\end{equation}
where the maximum is taken independently across the columns of $\mathcal{B}^\top \hwrench\in\mathbb{R}^{b\times (d c_{\mathrm{h},k})}$. We compute the robot support function $\rsupport$ in the same way.

Because every friction-cone edge carries unit-force, \Cref{eq:support_function} is the support function of $\mathrm{conv}\{w^{i,j}_{\mathrm{h},k}\}$, i.e.\ of the set of net wrenches achievable under a unit force budget \emph{shared}
across all contacts. We adopt a normalizing force budget because contact forces are not observable in our demonstrations, and absolute magnitudes transfer poorly across embodiments and from sim to real. This descriptor characterizes the directional capability of a contact set, and not a dynamically realizable wrench. It serves as similarity measure between the wrench geometry induced by human and robot grasp contacts. Actuation limits, hand kinematics, and object dynamics are enforced implicitly through RL in simulation.

We then define the contact wrench-space reward by comparing the robot support function $\rsupport$ against the human reference function $\hsupport$ with a relative tolerance $\beta$. For each object part $k$, the reward is defined as:
\begin{equation}
\label{eq:cws}
r_{\mathrm{cws}}^{k}
=
\exp\left(
-\frac{
\left\|
\max\left(0,(1-\beta)\hsupport-\rsupport\right)
\right\|_2^2
}{v_{\mathrm{cws}}}
-\frac{
\left\|
\max\left(0,\rsupport-(1+\beta)\hsupport\right)
\right\|_2^2
}{v_{\mathrm{cws}}}
\right),
\end{equation}
where the maximum is applied element-wise. The first term rewards robot contacts whose support is larger than the lower tolerance bound, while the second term rewards support values that do not exceed the upper tolerance bound. $v_{\mathrm{cws}}$ is the variance for the exponential kernel.

Since the exponential kernel in~\Cref{eq:cws} always yields positive rewards, we additionally penalize contact mismatches. In particular, we penalize unintended contacts $r_{\mathrm{unintend}}^k$ when $\hsupport=0$ but $\rsupport>0$, and missed contacts $r_{\mathrm{miss}}^k$ when $\hsupport>0$ but $\rsupport=0$. This prevents the policy from creating extra contacts not present in the human demonstration, or fails to reproduce contacts that are required by the reference.

\subsection{Efficient, Robust, and Generalizable Policy Learning}
\label{subsec:general}

In addition to the contact reward in~\Cref{subsec:contact_reward}, we incorporate several mechanisms to improve training efficiency and robustness. First, we reset the simulator to arbitrary states along the reference trajectory and keep the VOC fully active for a short stabilization window, allowing the robot to recover contact before assistance is annealed. Second, we perturb object parts with wrenches sampled from the human contact wrench matrix $\hwrench$, producing task-relevant disturbances aligned with the demonstrated contact mechanics. We also use a residual action space with retargeted robot motion as the prior~\cite{li2025maniptrans}, and anneal the VOC through a curriculum~\cite{mandi2026dexmachina}.

When contact estimates are noisy due to hand-object misalignment, as in demonstrations reconstructed from RGB videos, directly matching human contact wrenches can be unreliable. In such scenarios we pivot from direct matching to a reduced contact wrench guidance objective that rewards the robot for generating positive support along each wrench-space basis direction:
$r_{\mathrm{fc}}^{k}
= \frac{1}{B}\sum_{b=1}^{B}
\mathbbm{1}\!\left[\sigma_{r,k,b} > \epsilon \right]$, 
where $\sigma_{r,k,b}$ is the robot support value for object part $k$ along basis direction $b$, and $\epsilon$ is a small threshold. Maximizing this objective is equivalent to maximizing the the force closure objective.
% design an objective $r^{k}_{\mathrm{fc}}$ that densely rewards stable contact which when maximized is equivalent force closure. Concretely, given $B$ pre-sampled unit basis directions and the support function $\sigma_{r,k}$, we reward non-zero support along each basis direction greater than a threshold $\epsilon$ , normalized by the number of basis directions as,

\algabrvname{} naturally extends to whole-body manipulation with either \emph{hand-only} or \emph{whole-body} references. For hand-only references, e.g., egocentric reconstructions, we train an inpainting module~\cite{wang2026motionbricksscalablerealtimemotions} to predict full-body motion from end-effector trajectories, then apply the same RL formulation with contact wrench-space rewards. For whole-body references, e.g., third-person reconstructions, full-body motion is available, but finger reconstruction is often inaccurate, causing noisy hand-object interactions~\cite{xie2026cari4d}. We therefore train with the reduced force-closure objective $r_{\mathrm{fc}}^{k}$.

% CHORD directly extends to learning whole-body manipulation, with either \emph{hand-only} or \emph{whole-body} references. Given hand-only references, such as those obtained from egocentric reconstruction, we build on recent advances in whole-body inbetweening~\cite{wang2026motionbricksscalablerealtimemotions} and train an inpainting module to predict the whole body motion given end-effector trajectories. Utilizing the predicted whole-body motion, we then apply the same RL formulation with contact wrench space rewards to learn the task. While whole-body references such as those obtained from third-person view reconstruction directly provide the whole body motion, it is more common to have inaccurate hand finger reconstruction, resulting in noisy and unreliable hand-object interactions \cite{xie2026cari4d}. In these settings, we utilize the reduced contact objective $r_{\mathrm{fc}}^{k}$ for training.

\subsection{Dexterous Manipulation Benchmark with Human References}
\label{subsec:benchmark}

\begin{figure}[h]
    
    \centering
    \includegraphics[width=0.9\linewidth]{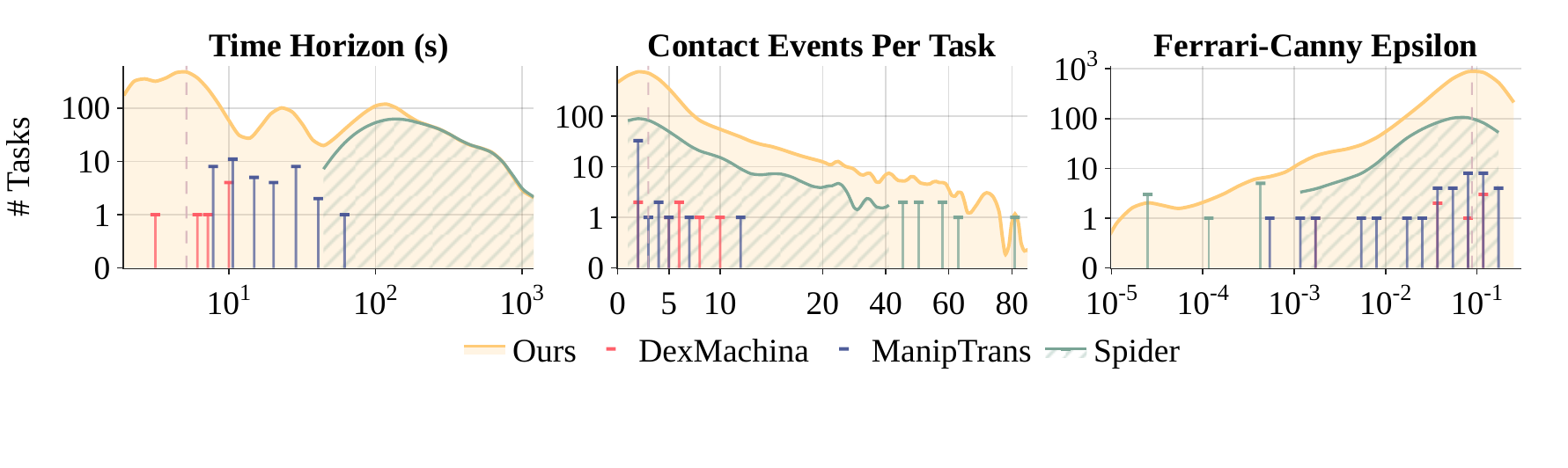}
    \caption{\textbf{Benchmark Distribution.} Our tasks span diverse, large-scale settings with long horizons and dense contact events, requiring dexterous contact strategies and coordinated manipulation.
    }
    \label{fig:task_lib}
    
\end{figure}

To further support the research on learning from human demonstrations and evaluate the proposed algorithm at a larger scale, we collect and process common human manipulation datasets~\cite{fan2023arctic,liu2024taco,oakink,hot3d,dexycb,grab,h2o} and import them into Isaac Lab for policy training~\cite{mittal2025isaaclab}. In total, we curate \numtask open-source tasks and convert them into simulatable, trainable environments. We also reconstruct additional demonstrations from in-house videos using hand-object reconstruction and tracking.
The resulting benchmark covers bimanual manipulation of single and multiple rigid objects, as well as articulated objects (\Cref{fig:teaser}). 
We evaluate task diversity against prior art~\cite{mandi2026dexmachina,li2025maniptrans,pan2025spider} across three core metrics: time horizon length, the number of contact events, and grasp stability, measured by the Ferrari-Canny metric \cite{ferrari1992planning}, as shown in~\Cref{fig:task_lib}.
These tasks require dexterous contact strategies beyond simple grasping and require coordination of both hands. See details in the Appendix.

% 196 (Arctic) + 373 (TACO) + 454 (Oakink) + 4037 (hot3d) + 372 (dexycb) + 304 (grab) + 3 (h2o) = 5739 (-1000 for safety)

\section{Experiments}
\label{sec:exp}

% We conduct experiments to evaluate the scalability of \algabrvname on large-scale tasks and compare it with baselines (\Cref{subsec:quant_analysis}); we further demonstrate its generalizability to whole-body manipulation (\Cref{subsec:exp_whole_body}) and real-world deployment (\Cref{subsec:real_world}).

\subsection{Quantitative Analysis on Dexterous Manipulation}
\label{subsec:quant_analysis}

% under diverse experimental settings. Due to differences in data collection hardware and processing pipelines, both the availability of information (e.g., gravity-aligned world frames, supporting surfaces, and object articulation) and the quality of that information (e.g., how reliably contact information can be recovered from human demonstrations) vary significantly across datasets. As a result, methods tailored to a particular dataset may perform well within that setting, but often fail to generalize to the broader range of challenges presented by other datasets.

% We apply a generic \algabrvname training recipe across a large-scale set of tasks spanning four open-source datasets: ARCTIC~\cite{fan2023arctic}, TACO~\cite{liu2024taco}, HOT3D~\cite{hot3d}, and OakInk2~\cite{oakink}. Collectively, these datasets cover rigid objects, articulated objects, and multi-object interactions, totaling \numtrainedtask tasks.

\begin{figure}[h]
    \centering
    \includegraphics[width=\linewidth]{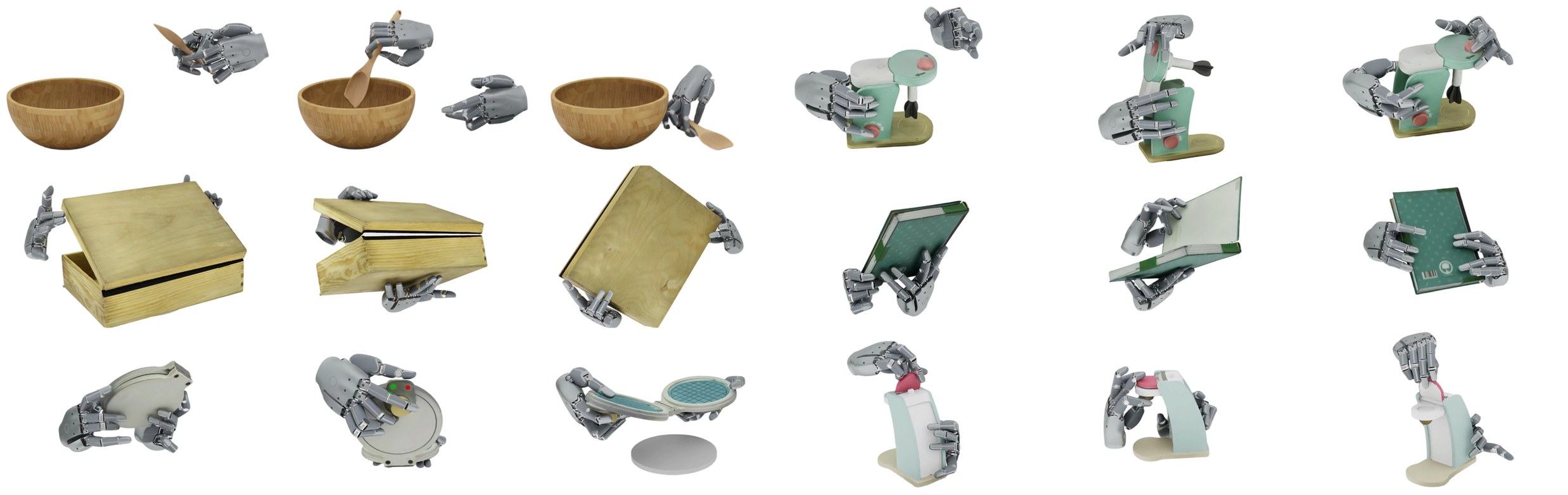}
    \caption{\textbf{Contact Rich Tasks:} The learned policies handle complex object interactions including picking, placing, handing over, stirring a bowl, and using articulations across objects of varying geometry.}
    \label{fig:sim}
    \vspace{-1em}
\end{figure}

\algabrvname uses FlashSAC~\cite{kim2026flashsac} with 4,096 environments, training each per-sequence policy on a single L40S GPU in ${\sim}$2 hours for bi-manual dexterous and whole-body manipulation tasks. This demonstrates training recipe scalability across heterogeneous tasks. We employ lightweight MLPs with \textit{no per-task hyperparameter tuning}. Full details are in the Appendix~\ref{app:rl}.

\paragraph{Evaluation on Large-Scale Tasks.}
We evaluate \algabrvname on \numtrainedtask tasks sampled from our benchmark~\Cref{subsec:benchmark}, using the same hyperparameters across all tasks, including VOC gains, curriculum schedules, and reward weights. These tasks were processed sequentially without filtering by difficulty, trainability, or success. To the best of our knowledge, \algabrvname is the first RL-based dexterous manipulation method evaluated at this scale. 

Qualitatively, \Cref{fig:sim} shows that \algabrvname solves diverse contact-rich tasks, including single-hand and bimanual pick-and-place, handover, coordinated manipulation, articulated-object manipulation, and tool use. The learned behaviors complete the tasks while closely following the human demonstrations. With only simple keypoint-based inverse kinematics, \algabrvname remains robust to retargeting noise and hand interpenetration by using wrench-space guidance.

\begin{figure}[h]
    \centering
    \includegraphics[width=0.95\linewidth]{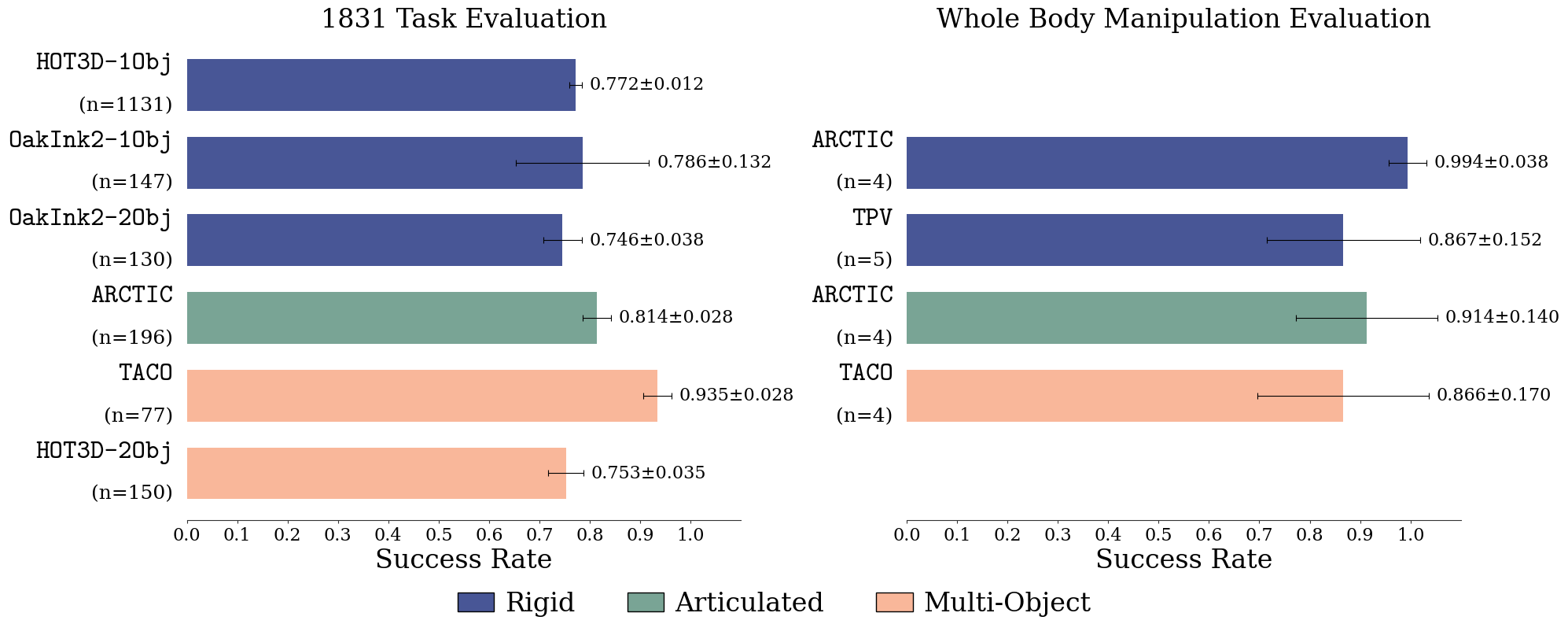}
    \caption{
    \algabrvname{} success rates on \textit{left:} \numtrainedtask manipulation tasks across four datasets, spanning both single-object and multi-object tasks involving rigid and articulated objects, and \textit{right:} 17 of these subtasks, fully grounded on a humanoid robot equipped with articulated hands.
    }
    \label{fig:scale-figure}
    
\end{figure}

For large-scale evaluation, we report task completion ratio as the primary metric. A rollout is considered successful if its position tracking error is below $15,\mathrm{cm}$ and its rotation tracking error is below $40^\circ$. We adopt this criterion because it empirically correlates strongly with human judgments of task success at large scale. Subsequent sections use standard metrics for fair comparisons.
\Cref{fig:scale-figure} (left) shows consistently strong performance across \numtrainedtask{} tasks spanning diverse object types and interaction horizons. We find that stable training requires the contact wrench reward to converge before VOC decay begins. Tasks involving small contact regions or delicate interactions generally require longer optimization.

\paragraph{Baselines Comparison.}
We compare \algabrvname against state-of-the-art dexterous manipulation methods that learn from human demonstrations, including ManipTrans (MT)~\cite{li2025maniptrans}, DexMachina (DM)~\cite{mandi2026dexmachina}, Spider (SP)~\cite{pan2025spider}, and Human2Sim2Robot (H2S2R)~\cite{pmlr-v305-lum25a}. Details of our baseline implementations and adaptations are provided in Appendix~\ref{app:baselineImplementation}.
% For each baseline, we evaluate it on the method's original task suite using its corresponding metrics with Sharpa hand. 
% After excluding tasks that cannot be reliably simulated in Isaac Lab due to asset-quality issues, we evaluate our method using 8 tasks from ManipTrans (\texttt{MT}), 3 from SPIDER (\texttt{SP}), and all 7 from DexMachina (\texttt{DM}).
% We also select 9 tasks from our benchmark, covering single-object manipulation (\texttt{Ours-1}) and two-object interaction tasks (\texttt{Ours-2}). 
% We report each method using the metrics from the corresponding baseline: area under the object-tracking-error curve (\texttt{AUC}) from DexMachina~\cite{mandi2026dexmachina}, success rate from SPIDER (\texttt{SP-SR})~\cite{pan2025spider}, and success rate from ManipTrans (\texttt{MT-SR})~\cite{li2025maniptrans}. 
We adopted standardized metrics ($\texttt{AUC}$ from DexMachina, $\texttt{SP-SR}$ from Spider, $\texttt{MP-SR}$ from ManipTrans). In addition, we also report functional continuous error metrics: articulated joint position error ($\texttt{JPE}$, $^\circ$), multi-object relative position error ($\texttt{RPE}$, $\text{cm}$), and keypoint mean per-frame pose error ($\texttt{MPPE}$, $\text{cm}$). We selected 60 tasks from the benchmark uniformly to cover rigid, articulated, and multi-rigid manipulation with diverse horizons.

\begin{table}[t]
\centering
\small
\renewcommand{\arraystretch}{1.25}
\setlength{\tabcolsep}{6pt}

\begin{tabular*}{\textwidth}{@{\extracolsep{\fill}}lcccccc@{}}
\toprule
\textbf{Method}
& AUC $\uparrow$
& SP-SR $\uparrow$
& MP-SR $\uparrow$
& JPE ($^\circ$) $\downarrow$
& RPE (cm) $\downarrow$
& MPPE (cm) $\downarrow$ \\
\midrule
DM
& $0.737 \pm 0.053$
& $0.690 \pm 0.068$
& $0.569 \pm 0.097$
& $13.10 \pm 2.31$
& $9.92 \pm 1.64$
& $15.88 \pm 0.70$ \\

MT
& $0.506 \pm 0.006$
& $0.423 \pm 0.011$
& $0.323 \pm 0.003$
& $30.07 \pm 0.61$
& $9.91 \pm 0.42$
& $21.24 \pm 6.19$ \\

SP
& $0.804 \pm 0.044$
& $0.911 \pm 0.039$
& $0.789 \pm 0.047$
& $41.35 \pm 2.44$
& $6.99 \pm 1.56$
& $14.28 \pm 2.79$ \\

H2S2R
& $0.730 \pm 0.056$
& $0.686 \pm 0.068$
& $0.569 \pm 0.085$
& $11.80 \pm 1.50$
& $9.78 \pm 1.03$
& $16.59 \pm 0.51$ \\

\textbf{CHORD}
& $\mathbf{0.918 \pm 0.018}$
& $\mathbf{0.926 \pm 0.022}$
& $\mathbf{0.894 \pm 0.032}$
& $\mathbf{4.34 \pm 0.25}$
& $\mathbf{5.14 \pm 0.54}$
& $\mathbf{5.11 \pm 0.39}$ \\

\midrule

Pos. Only
& $0.862 \pm 0.019$
& $0.856 \pm 0.028$
& $0.780 \pm 0.042$
& $6.41 \pm 0.44$
& $7.81 \pm 0.64$
& $7.63 \pm 0.06$ \\

No Cont.
& $0.774 \pm 0.036$
& $0.756 \pm 0.059$
& $0.612 \pm 0.055$
& $9.81 \pm 0.73$
& $10.91 \pm 1.24$
& $14.38 \pm 2.23$ \\

\bottomrule
\end{tabular*}
\caption{\textbf{Baseline Comparison and Reward Ablation.}
Comparison with prior methods and ablation of reward formulations across 60 tasks.
Results are reported as mean $\pm$ standard deviation over three training seeds,
with 4,096 environments per seed. Best results are highlighted in bold.}
\label{tab:comparison_prior_work}
\end{table}

Table~\ref{tab:comparison_prior_work} shows that \algabrvname achieves high success rates on both the baseline suites and our task suite, consistently matching or outperforming state-of-the-art methods under their original metrics. These results agree with the large-scale evaluation in~\Cref{fig:scale-figure}, showing that \algabrvname provides both broad manipulation capability and the precision needed for task completion.
Moreover, prior methods are limited to subsets of rigid-object, articulated-object, or multi-object manipulation, whereas \algabrvname succeeds across all three categories. This broader coverage is important for building manipulation capabilities that are robust across task types.
% Additional robustness analysis and ablations are provided in the Appendix.

\paragraph{Contact Reward Ablations.}
% We compare three reward formulations with progressively reduced contact guidance: \algabrvname (contact wrench support reward), Position Only (contact position reward, as in DexMachina), and No Contact (task and imitation rewards only). Position Only evaluates the benefit of wrench guidance beyond contact position, while No Contact evaluates the contribution of contact guidance overall.
% All variants share identical non-contact rewards for object pose tracking, hand keypoint tracking, and hand joint position tracking. 
% As shown in Table \ref{tab:comparison_prior_work}, performance improves with richer contact guidance. \algabrvname achieves the best overall results, demonstrating the benefit of supervision in contact wrench space. Replacing wrench guidance with contact position rewards (Position Only) reduces performance, suggesting that matching contact locations alone does not adequately capture the interaction dynamics required for contact-rich manipulation. 

We compare three reward formulations: \algabrvname (contact wrench support), Position Only (contact position~\cite{mandi2026dexmachina}), and No Contact (task and imitation rewards only). All variants share identical object pose, hand keypoint, and hand joint position tracking rewards.
Performance improves with richer contact guidance (Table~\ref{tab:comparison_prior_work}): \algabrvname performs best, followed by Position Only and No Contact. These results support the benefit of explicit contact guidance and suggest that contact locations alone do not fully capture the dynamics of contact-rich manipulation. No Contact performs worst overall, highlighting the limitations of relying solely on task and imitation rewards without explicit contact guidance.

% For example, ManipTrans and Spider do not explicitly model articulated objects. Consequently, tasks involving articulated structures, such as opening a microwave door, are often solved by exploiting object dynamics through contact forces rather than by performing the intended articulation manipulation. In contrast, \algabrvname is able to naturally handle articulated interactions while maintaining strong performance across rigid, articulated, and multi-object scenarios.

\begin{figure}[H]
    \centering
    \includegraphics[width=0.85\linewidth]{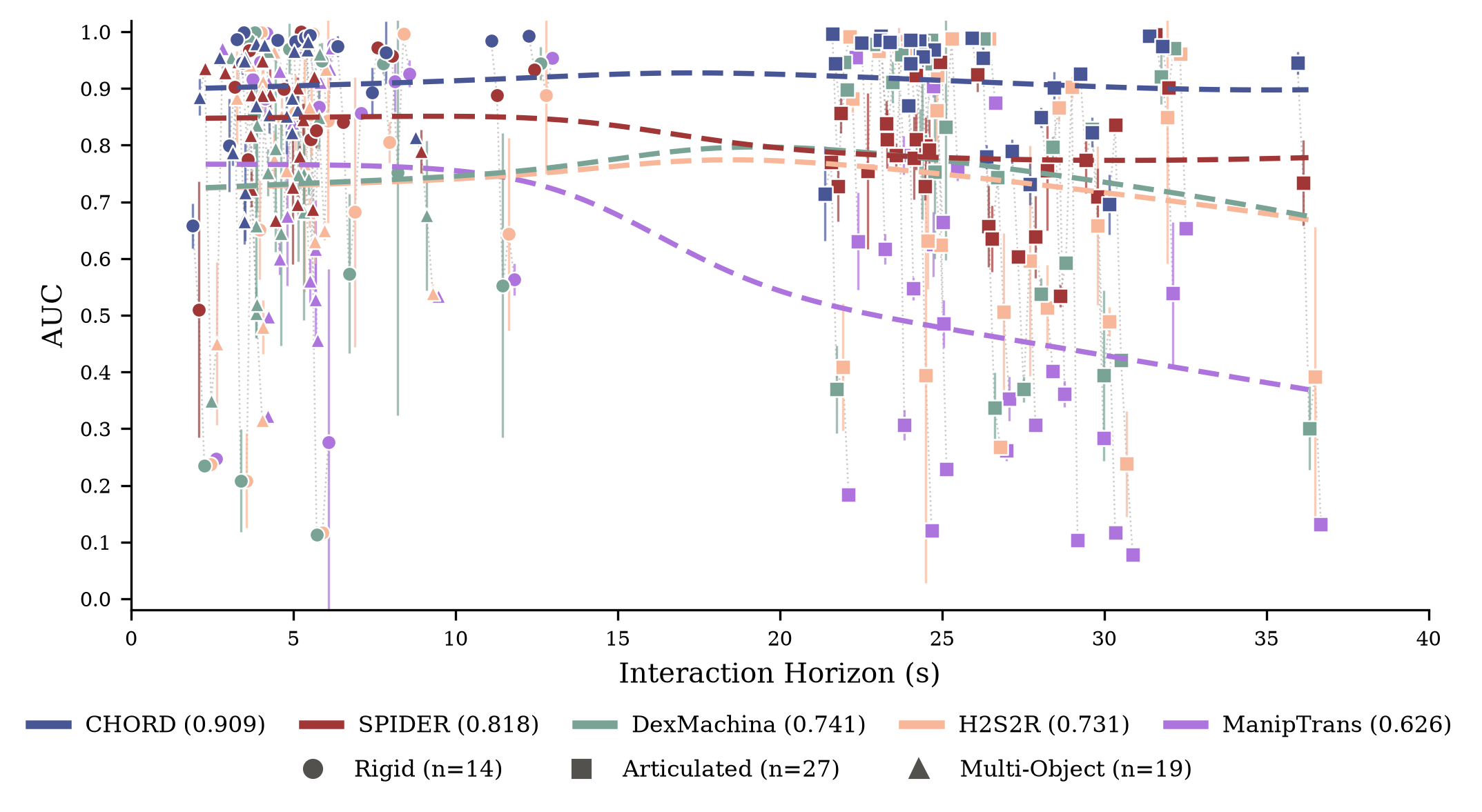}
    \caption{\textbf{\algabrvname maintains high tracking accuracy over long horizons.}
Object-tracking AUC versus interaction horizon for 60 sequences.
Colors denote methods; marker shapes denote sequence types.
Markers show means across seeds, with error bars indicating standard deviations. Dashed lines show each method's trend.
}
\label{fig:horizon_auc}
\end{figure}

\paragraph{Diverse Capabilities Enable Long-Horizon Manipulation.}
% Contact wrench serves as a unified abstraction that enables \algabrvname to perform a diverse range of dexterous manipulation tasks. This diversity is a key enabler of long-horizon task execution: by composing a broad repertoire of manipulation behaviors through a single mechanism, \algabrvname scales to task sequences spanning unprecedented temporal horizons. \Cref{fig:horizon_auc} presents performance measured using the AUC metric. The evaluated sequences cover a broad spectrum of interaction horizons, ranging from short manipulation episodes to extended task sequences lasting nearly a minute. Across this range, \algabrvname consistently achieves strong performance. For most sequences, it maintains near-saturated tracking accuracy (AUC $\approx$ 0.85$-$0.98) even on the longest horizons, whereas baseline methods exhibit substantial degradation as the interaction horizon increases.
Contact wrench serves as a unified abstraction for diverse dexterous manipulation tasks. By guiding these behaviors through a shared mechanism, \algabrvname supports longer manipulation sequences. \Cref{fig:horizon_auc} reports object-tracking AUC across sequences ranging from short manipulation episodes to extended interactions. Across this range, \algabrvname maintains high tracking accuracy for most sequences (AUC $\approx 0.85$--$0.98$), including those with the longest horizons, while baseline methods degrade substantially as the interaction horizon increases.

\subsection{Evaluation on Whole-Body Manipulation}
\label{subsec:exp_whole_body}
\begin{figure}
    \centering
    \includegraphics[width=\linewidth]{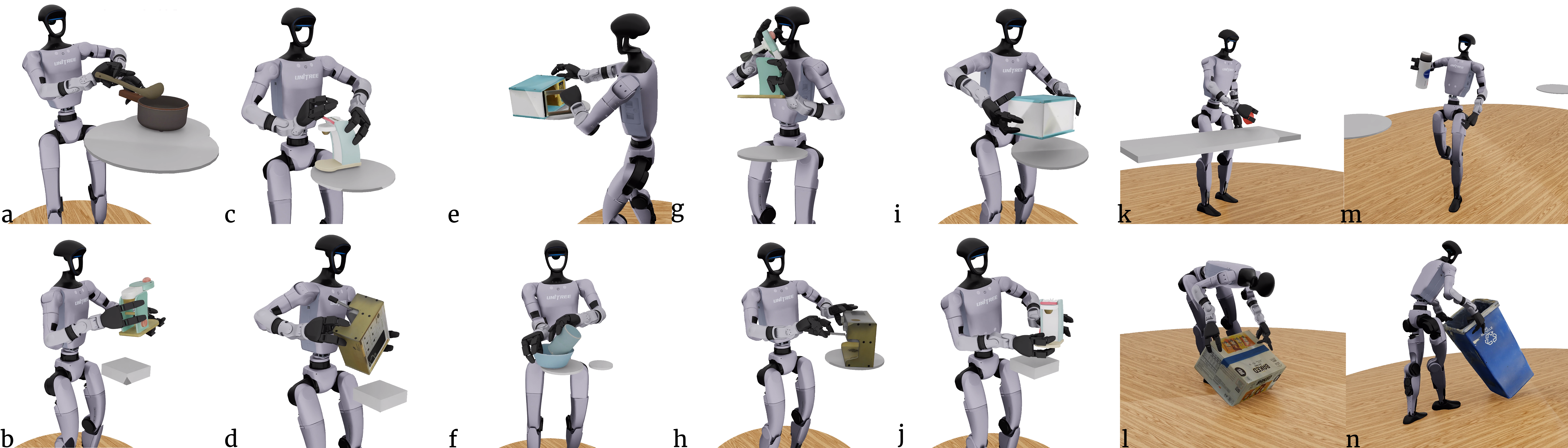}
    \caption{\textbf{Whole-body manipulation results.}
(a--j) Results using hand-only references, where full-body motion is completed by an inpainting module and RL training. 
(k--n) Results using whole-body references, where the reduced force-closure objective is used during RL training.}
    \label{fig:whole_body_sim}
    \vspace{-1em}
\end{figure}

We evaluate the whole-body extension of \algabrvname described in~\Cref{subsec:general} on representative tasks from both hand-only and whole-body references. The hand-only references include 12 tasks: 4 rigid-object manipulation, 4 articulated-object manipulation, and 4 multi-object manipulation tasks. The whole-body references are reconstructed from third-person-view (\texttt{TPV}) videos and include 5 loco-manipulation tasks (\Cref{fig:whole_body_sim}).
The results in~\Cref{fig:scale-figure} (right) show that \algabrvname generalizes across embodiments, successfully transferring hand-only demonstrations to the G1 robot with Dex3 three-fingered hands under whole-body kinematic constraints. Wrench-space alignment enables the policy to learn effective contact behaviors despite the morphology difference between the human hand and the robot hand. For the $\texttt{TPV}$ tasks~\cite{xie2026cari4d}, hand-object reconstruction noise makes the reference contact wrench unreliable, so we instead use the force-closure objective (\Cref{subsec:general}), which achieves consistent success across all $\texttt{TPV}$ tasks. See Appendix~\ref{app:whole_body} for details.

% Our success rate on the tasks generated from hand-only references reported in~\Cref{fig:whole_body_sim} demonstrates the generalizability and cross-embodiment capability of CHORD, where utilizing wrench space alignment enables us to robustly learn successful behaviors on the Dex3 three-fingered hand with the G1 kinematic constraints.
% On tasks reconstructed from third-person view~\cite{xie2026cari4d}, the contact wrench reward is not reliable due to noisy hand object reconstruction and we utilize the force closure objective in~\Cref{subsec:general} instead.
% We find that this objective results in consistent success across all five TPV tasks despite not having a reference wrench to align to, as seen in Fig. \ref{fig:whole_body_sim}.

% While contact position rewards would suffer \bowen{do we have results to support this?} when transferring to a different embodiment due to the mismatch in embodiment constraints, wrench space alignment guides learning towards contact behaviors that reproduce the induced object motion independently of embodiment constraints.

\begin{figure}[h]
    \centering
    \includegraphics[width=\linewidth]{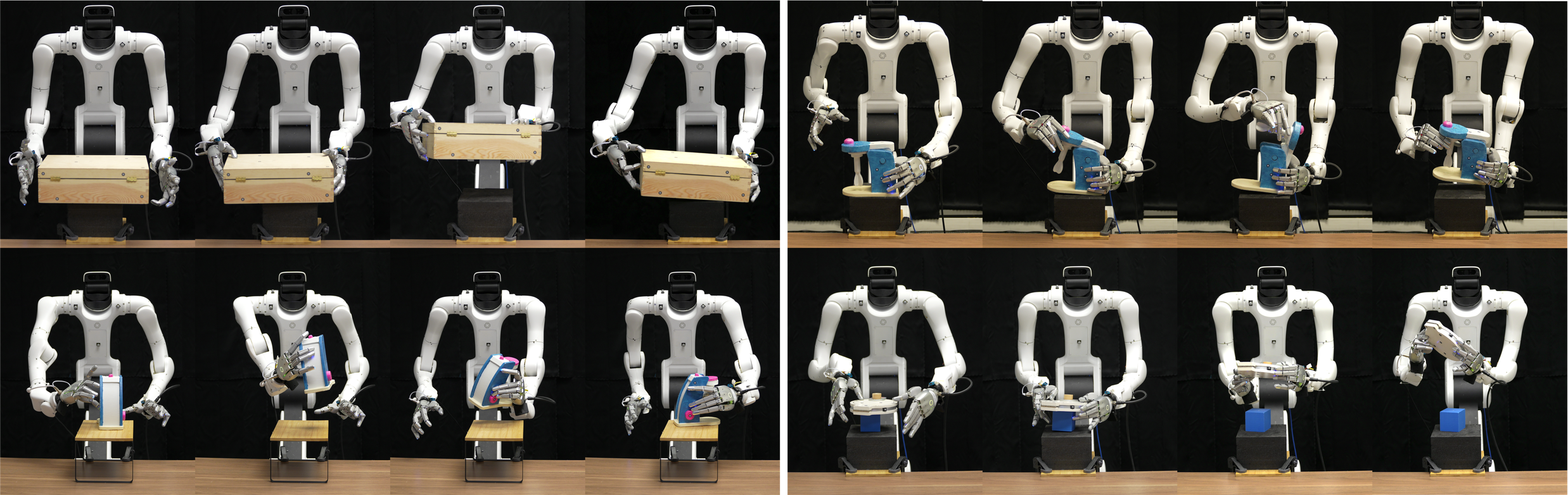}
    \caption{
    \textbf{Real-World.} The top-left shows closed-loop, and the others show open-loop deployment.
}
    \label{fig:real_world}
    \vspace{-1.23em}
\end{figure}

\subsection{Evaluation on Real World}
\label{subsec:real_world}
% To evaluate real-world transfer, we deploy the learned policies on a Dexmate robot with two Sharpa dexterous hands. 
% We evaluate two deployment modes: In open-loop deployment, we use action-chunk execution~\cite{li2025maniptrans, pan2025spider} . In closed-loop deployment, the robot wrist, finger tip pose, and object pose are tracked using a motion-capture system \Cref{app:vicon}. At each timestep, this feedback together with robot proprioceptive state is combined into an observation tensor that is fed to the policy to predict the next residual action that, together with the reference trajectory, determines the robot command. As shown in~\Cref{fig:real_world}, the policies successfully manipulate both rigid and articulated objects with bimanual coordination, requiring accurate contact control. % TODO: expand on real robot results

To evaluate real-world transfer, we deploy the learned policies on a Dexmate robot equipped with two Sharpa dexterous hands. A motion-capture system tracks robot and object poses. We evaluate both open-loop execution of actions ~\cite{li2025maniptrans, pan2025spider} and closed-loop policy inference. As shown in~\Cref{fig:real_world}, the policies successfully manipulate rigid and articulated objects in tasks requiring bimanual coordination and precise contact control.
\algabrvname succeeds in 5/5 trials for bowl handover, 3/5 for in-hand plate reorientation, 5/5 for bowl--plate stacking, 5/5 for mixer opening and closing, and 5/5 for alternating pick-and-place with a capsule machine. 
% Observed failures arise from sim-to-real discrepancies in initial object placement and contact transitions during open-loop execution.
On box pick-and-place, open-loop execution succeeds in 6/7 trials, compared with 7/7 for closed-loop inference. 
% Here, closed-loop execution refers to using live robot and object state feedback to correct actions while tracking a fixed demonstrated reference trajectory. The policy does not perform online replanning of that reference.
See Appendix~\ref{app:hardware} for details.

\section{Conclusion and Discussion}
\label{sec:conclusion}

We present \algabrvname, a framework that uses contact wrench-space
guidance to learn dexterous manipulation from human demonstrations.
By comparing the wrench capabilities of human and robot contacts,
\algabrvname accommodates differences in hand morphology and contact
geometry. We introduce a benchmark of \numtask manipulation tasks
and evaluate \algabrvname on \numtrainedtask tasks, achieving a
success rate of \successrate. Whole-body experiments and real-world
deployment through both open-loop execution and closed-loop control
further demonstrate the applicability of the formulation.

\section{Limitations}
\label{sec:limitation}

% All submissions should include a Limitations section (counted toward the 8-page limit), explicitly describing limiting assumptions, failure modes, and other limitations of the results and experiments, and how these might be addressed in the future.

Despite the scalability and robustness of \algabrvname, several limitations remain. First, real-world deployment currently relies on state-based observations. Second, effective contact guidance requires relatively clean human demonstrations, as noisy contacts can degrade reward quality and will require a force-closure assumption. Third, exact object placement may not be essential in some tasks, while a small pose error can lead to functional failure in others.
% Future work should address vision-based deployment, robustness to noisier demonstrations, and more task-aware success metrics.

\section{Acknowledgments}
\label{sec:acknowledge}

We thank Chao-Yeh Chen, Lin Duan, Qingsi Wang, Xiufeng Xie, Yu-Hsiang Huang, Tiffany Chen, and David Chu for their help in standing up Dyn-HaMR. We are grateful to Ziqiang Huang, Jingrui Yu, and Chirag Majithia for their contributions to object reconstruction. We thank Daniel Zou, Jazmin Sanchez, and Hesam Rabeti for their support with third-person-view data collection, processing, and reconstruction, and Xianghui Xie and Stan Birchfield for their guidance on third-person-view reconstruction. We are also grateful to Edy Susanto Lim, Ehsan Hassani, Sam Wu, Jun Saito, Michael De Ruyter, Jiefeng Li, Ye Yuan, Umar Iqbal, and Simon Yuen for their advice on the human body model. We thank Jie Xu and Yashraj Narang for discussions on contact and simulation. Finally, we thank Spencer Huang and Rev Lebaredian for their leadership and guidance.

\bibliographystyle{unsrt}
\bibliography{references}  % References your paper.bib file (without .bib extension)

\clearpage
\appendix

\begin{center}
    {\LARGE\bfseries Appendix
} \\ [1.5em]
\website \\
\end{center}

This chapter supplements our manuscript, \emph{Learning Dexterous Manipulation Using Contact Wrench Guidance From Human Demonstration}. It provides additional details on:
\begin{itemize}[leftmargin=1.5em,noitemsep,topsep=0pt]
\item Studies on the correlation of our proposed contact reward to task success (\Cref{app:contact_reward_correlation})
\item Ablation studies on noise in demonstrations (\Cref{app:noise_ablation}) and contact wrench parameterization (\Cref{app:wrench_param})
\item Reinforcement learning implementation details (\Cref{app:rl})
\item Reconstruction and learning from human video (\Cref{app:reconstruction})
\item Benchmark details and statistics (\Cref{app:task_library})
\item Whole-body motion planning and control (\Cref{app:whole_body})
\item Hardware experiment setup (\Cref{app:hardware})
\item Comparison of \algabrvname{} with teleoperation (\Cref{app:Teleop_VS_RL})
\item Baseline implementation details (\Cref{app:baselineImplementation})
\end{itemize}

\section{Correlation between Contact Wrench Reward and Task Success}
\label{app:contact_reward_correlation}

\begin{figure}[h]
    
    \centering
    \includegraphics[width=0.55\linewidth]{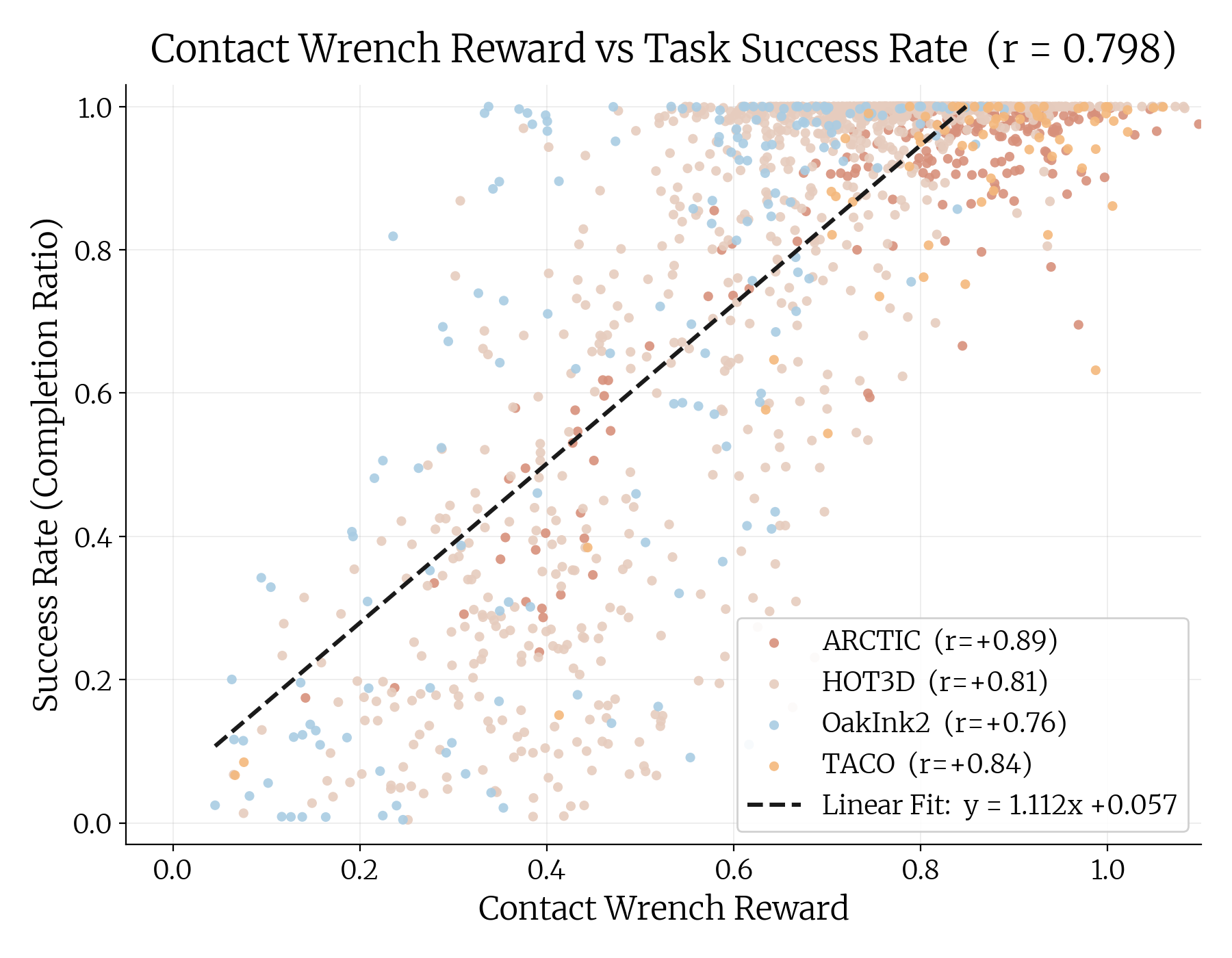}
    \caption{Correlation between contact wrench reward and task success rate.
    }
    \label{fig:contact_success_correlation}
    \vspace{-0.7em}
\end{figure}

To assess whether the contact-wrench-support (CWS) reward is predictive of downstream task performance, we aggregated \numtrainedtask runs in our benchmark. For each run, we extracted the CWS reward and the task success (evaluation completion ratio) over \numtrainedtask tasks. The achieved CWS reward was normalized on a per-sequence basis to render magnitudes comparable across datasets of differing contact density and trajectory length in~\Cref{fig:contact_success_correlation}.

Across all runs, we observe a strong positive association between the normalized CWS reward and task success (Pearson $r \approx 0.80$), which holds consistently within every dataset ($r = 0.76$--$0.89$). The relationship is monotonic but saturating: success rises steeply with CWS reward and then plateaus near unity, so that an ordinary least squares line---reported here as a first-order summary---captures the overall trend (explaining roughly two-thirds of the variance) while necessarily understating the fit in the high-reward regime. These results indicate that the degree to which a policy satisfies the contact-wrench-support objective is a reliable correlate of its manipulation success, supporting CWS reward as a useful training signal and proxy metric.

\section{Extended Ablation Studies}
\label{app:ablation}

To understand the effectiveness of the contact wrench reward and the robustness of \algabrvname, we conduct two ablation studies. We select two representative sequences that are long-horizon and involve articulated object manipulation capabilities from ARCTIC: $\texttt{box grab}$, $\texttt{mixer use}$.
% spanning a range of task horizons, manipulation complexities, and contact-rich interaction types from three datasets: ARCTIC, Hot3D, and TACO. Specifically, the selected sequences are:
% \begin{itemize}[leftmargin=1.5em,noitemsep,topsep=0pt]
% \item \textbf{Single-object grasping:} \texttt{ARCTIC-box-grab} (983 frames)
% \item \textbf{Articulated object manipulation:} \texttt{ARCTIC-mixer-use} (856 frames)
% \item \textbf{Two-object manipulation:}
% \begin{itemize}[leftmargin=1.5em,noitemsep,topsep=0pt]
% \item \texttt{TACO-kettle-kettle} (382 frames), \texttt{Hot3D-pour-from-moka-to-cup} (210 frames), and \texttt{Hot3D-put-bowl-on-plate} (150 frames)
% \end{itemize}
% \end{itemize}
% Specifically, the selected sequences consist of a single-object grasping task, \texttt{ARCTIC-box-grab} (983 frames); an articulated object manipulation task, \texttt{ARCTIC-mixer-use} (856 frames); and three two-object manipulation tasks: \texttt{TACO-kettle-kettle} (382 frames), \texttt{Hot3D-pour-from-moka-to-cup} (210 frames), and \texttt{Hot3D-put-bowl-on-plate} (150 frames).
Within each ablation study, all studies are evaluated using a set of three random seeds.

\subsection{Robustness on Human Demonstration Noise}
\label{app:noise_ablation}

To evaluate the robustness of \algabrvname{}, we conduct an ablation study in which increasing levels of noise are injected into the demonstration data. The goal is to understand how sensitive \algabrvname is to imperfections commonly encountered in reconstructed demonstrations. We sample random walk noise as follows,
\begin{table}[h]
\centering
\label{tab:noise_bounds}
\begin{tabular}{c|cc|cc}
\toprule
Scale & wrist pos. (mm) & wrist ori. (deg) & contact pos. (mm) & contact normal (deg) \\
\midrule
1 & 10 & 5 & 5 & 5 \\
2 & 20 & 10 & 10 & 10 \\
3 & 30 & 15 & 15 & 15 \\
4 & 40 & 20 & 20 & 20 \\
5 & 50 & 25 & 25 & 25 \\
6 & 60 & 30 & 30 & 30 \\
% 7 & 70 & 35 & 35 & 35 \\
% 8 & 80 & 40 & 40 & 40 \\
\bottomrule
\end{tabular}
\caption{Noise bounds per scale.}
\end{table}

\begin{figure}[H]
    
    \centering
    \includegraphics[width=0.65\linewidth]{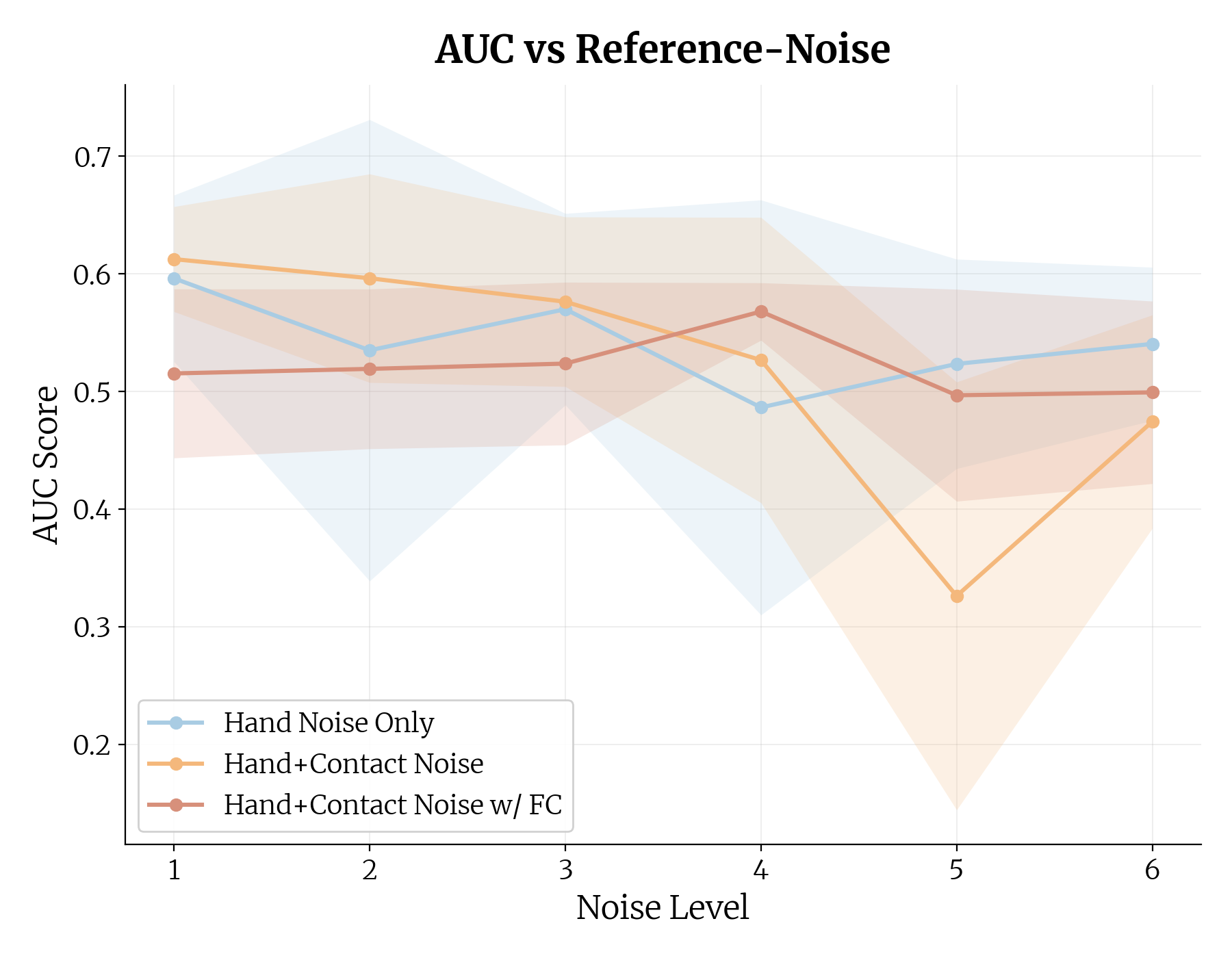}
    \caption{Learning performance vs demonstration noise level.
    }
    \label{fig:noise_ablation}
\end{figure}

We consider three settings:

\begin{itemize}[leftmargin=1.5em,noitemsep,topsep=0pt]
\item \textbf{Noisy Hand Pose + Contact Wrench Reward} ($\texttt{Hand Noise Only}$): noise is injected into the hand pose, while contact guidance remains noise-free. Contact supervision is provided using the contact-wrench-space reward. This setting simulates human demonstrations captured through video-based motion reconstruction, where contact information may still be obtained from tactile gloves.

\item \textbf{Noisy Hand Pose + Noisy Contact + Contact Wrench Reward} ($\texttt{Hand+Contact Noise}$): noise is injected into both the hand pose and contact guidance, including contact positions and surface normals, thereby perturbing the wrench supports. Contact supervision is provided using the contact-wrench reward. This setting simulates human demonstrations captured purely from video-based reconstruction, without additional sensors.

\item \textbf{Noisy Hand Pose + Noisy Contact + Force-Closure Reward} ($\texttt{Hand+Contact Noise w/ FC}$): noise is injected into both the hand pose and contact guidance, including contact positions and surface normals. However, training is guided by a force-closure reward rather than the contact-wrench reward. This setting evaluates whether wrench-space supervision is more robust to noisy demonstrations than force-closure-based objectives.
\end{itemize}

\Cref{fig:noise_ablation} compares task performance as the magnitude of injected noise increases. \algabrvname{} remains robust to moderate hand-pose noise, with gradual performance degradation, suggesting that the method can tolerate realistic levels of state estimation and tracking error.

When noise is additionally introduced into contact guidance, performance decreases more noticeably, reflecting the increased difficulty of recovering accurate contact interactions from corrupted demonstration data. At very high noise levels, force-closure guidance can outperform noisy wrench-space guidance, since heavily corrupted wrench supports may provide misleading supervision. However, we also observe that relying on force-closure guidance reduces dexterity, as it prioritizes stable grasps over faithful imitation of human manipulation behavior.

\subsection{Performance on Wrench Parameterizations}
\label{app:wrench_param}

\begin{table}[t]
\centering
\renewcommand{\arraystretch}{1.25}
\begin{tabular}{c c c c c c}
\toprule
$b$ & $d$ & $\mu$ & \texttt{AUC} & \texttt{SP-SR} & \texttt{MP-SR} \\
\midrule
\textbf{512} & \textbf{8} & \textbf{0.1}
    & \textbf{0.875 $\pm$ 0.017}
    & 0.895 $\pm$ 0.023
    & \textbf{0.824 $\pm$ 0.032} \\
256 & 8 & 0.1
    & 0.864 $\pm$ 0.004
    & 0.883 $\pm$ 0.010
    & 0.795 $\pm$ 0.004 \\
1024 & 8 & 0.1
    & 0.867 $\pm$ 0.012
    & 0.883 $\pm$ 0.027
    & 0.803 $\pm$ 0.026 \\
512 & 4 & 0.1
    & 0.867 $\pm$ 0.005
    & \textbf{0.901 $\pm$ 0.024}
    & 0.803 $\pm$ 0.004 \\
512 & 8 & 0.5
    & 0.861 $\pm$ 0.002
    & 0.870 $\pm$ 0.010
    & 0.797 $\pm$ 0.007 \\
512 & 8 & 1.0
    & 0.848 $\pm$ 0.030
    & 0.864 $\pm$ 0.039
    & 0.782 $\pm$ 0.051 \\
\bottomrule
\end{tabular}
\caption{Wrench ablation ($b$: 6-D support directions; $d$: friction-cone edges; $\mu$: friction coefficient).}
\label{tab:wrench_ablation}
\end{table}

Table~\ref{tab:wrench_ablation} confirms CHORD’s robustness to wrench parameters ($b, d, \mu$) across 20 tasks sampled from the set of 60, with 3 seeds; all submission experiments used $b=512$ (uniformly sampled directions on the unit sphere), $d=8$, and $\mu=0.1$ for computation efficiency and performance.

\section{Implementation Details of Reinforcement Learning}
\label{app:rl}

\paragraph{Observation Space.}
The policy observation is formed by concatenating proprioceptive state, object state, task command, action history, and contact information. Specifically, the observation includes the following terms:
\begin{itemize}[leftmargin=1.5em,noitemsep,topsep=0pt]
\item \textbf{Wrist state:} right and left wrist positions, orientations, and body-frame velocities.
\item \textbf{Finger state:} right and left finger joint positions and velocities, with joint positions scaled by joint limits.
\item \textbf{Object state:} flattened object/body positions and orientations for all objects in the scene.
\item \textbf{Command:} target deltas for wrist poses, finger joints, and object poses.
\item \textbf{Action history:} the previous raw policy action, together with processed right- and left-hand actions.
\item \textbf{Contact observation:} contact positions and force directions transformed into the wrist frame, with inactive contacts zeroed out.
\end{itemize}
Although noise configurations are defined for several observation terms, observation corruption is disabled in our experiments during large-scale experiments, but enabled in real-world closed-loop policy training.

\paragraph{Action Space.}
The policy controls residual actions for both hands. For each hand, the action consists of a 3D wrist position residual, a 3D wrist orientation residual, and finger joint residuals. The action scales are set to $0.05m$ for wrist position, $0.15 rad$ for wrist orientation, and $0.15rad$ for finger joints. We apply an exponential moving average with factor $0.3$ to smooth the actions. The resulting targets are executed through a wrist force/torque tracking controller and finger joint position targets.

For object control, the scene setup additionally creates one virtual object-control term for each rigid or articulated object. These terms have zero policy action dimension and therefore are not directly controlled by the policy. Instead, they apply curriculum-scaled PD wrenches or efforts internally.

\paragraph{Termination and Reset.}
Episodes terminate when any of the following conditions is satisfied:
\begin{itemize}[leftmargin=1.5em,noitemsep,topsep=0pt]
\item \textbf{Time-out:} the timestep counter reaches the retargeted motion horizon.
\item \textbf{Wrist deviation:} either wrist deviates too far from the commanded wrist trajectory.
\item \textbf{Object deviation:} any object exceeds the allowed position or orientation error threshold.
\end{itemize}

Each episode starts from a random frame along the retargeted motion. At reset, the environment restores the object root pose and velocity, resets articulated object joints when present, restores the right and left wrist root poses and velocities, and initializes the right and left finger joints to a partially open version of the reference pose using a random interpolation factor in $[0, 0.7]$. The per-environment virtual object-control scale is reset to $1.0$, followed by a $20$-step warm-up period before applying the curriculum target. Finally, the action terms clear raw actions, previous actions, wrist forces and torques, and finger targets.

\paragraph{Relative Task Reward.}
For two-object manipulation sequences, tracking each object independently is not always sufficient to capture the task-relevant geometry between objects. Many interactions, such as inserting,
pouring, scooping, or tool use, depend primarily on the relative pose between the manipulated objects rather than their absolute poses in the world. We therefore introduce a relative
object pose reward, $r_{\mathrm{relative}}$, that encourages the policy to reproduce the demonstrated spatial relationship between objects, defined as an exponential tracking kernel over both translational and rotational error between the two objects:
\[
r_{\mathrm{relative}}(t)
=
m(t)
\exp\left(-\frac{e_p(t)\cdot e_R(t)}{\textrm{var}_{\mathrm{rel}}}\right)
\]
where $\textrm{var}_{\mathrm{rel}}$ controls the tolerance of the translational and rotational components, respectively. The relative translation and rotation errors are defined as
$e_p(t) =
\left\|p_{1|0}(t) - \bar{p}_{1|0}(t)\right\|_2,
e_R(t) =
d_{\mathrm{geo}}\left(R_{1|0}(t), \bar{R}_{1|0}(t)\right),
$
where \(d_{\mathrm{geo}}\) is the geodesic rotation distance. 
The binary mask \(m(t)\) activates the reward only during demonstrated object-object interaction phases. 
Importantly, this gating depends solely on the demonstration inter-object distance, rather than the policy’s current object distance. This design prevents the policy from trivially avoiding the reward by keeping objects far apart, while still avoiding unnecessary relative-pose constraints during approach phases in which the demonstrated objects are not yet interacting.
For single-object sequences, the reward is zero. For multi-object sequences, the demonstrated relative poses and inter-object distances are precomputed once from the reference trajectory, and the reward is evaluated at every simulation step using the current object root poses. The multiplicative formulation ensures that high reward is achieved only when both the relative translation and relative orientation match the demonstration, making the reward directly aligned with preserving task-relevant object-object geometry.

\section{\algabrvname Validation from Human Video}
\label{app:reconstruction}

To validate CHORD end-to-end from raw egocentric human video, we select a brush
manipulation task. We reconstruct the hand--object interaction with a pipeline that
recovers, from a single monocular RGB video and a text prompt naming the object, a
per-frame MANO~\cite{romero2017embodied} hand pose and a textured object mesh with its 6-DoF pose,
expressed in a shared, roughly-metric space. This trajectory is the input we pass to CHORD
for retargeting and grounding into simulation.

Monocular depth from MoGe~\cite{wang2025moge} is the metric anchor for the pipeline: object scale,
hand scale and translation, and the camera trajectory are all aligned to the same depth,
placing the two streams in a common frame with a consistent hand--object displacement.
Each stream is otherwise estimated independently, until a final Gaussian-splat refinement
jointly optimizes and temporally smooths all outputs.

\paragraph{Object.}
We detect the object on a reference frame with an open-vocabulary detector (Grounding
DINO~\cite{liu2024groundingdino}) prompted by the text label, segment it across the video
(SAM2~\cite{ravi2024sam2}), and reconstruct a textured mesh from the reference frame
(SAM3D~\cite{sam3d2025}). As the mesh is recovered only up to scale, we estimate its metric
size by registering it against the MoGe depth, then track its 6-DoF pose across the video
(FoundationPose~\cite{wen2024foundationpose}).

\paragraph{Hand.}
We recover per-frame MANO parameters with a monocular hand tracker (ViPE~\cite{huang2025vipe} +
Dyn-HaMR~\cite{yu2025dynhamr}) and align them to metric depth: we render the hand and shift it
along its viewing ray so that its rendered depth matches MoGe, recovering a per-frame
translation and a global hand scale. The hand and object then share the same depth anchor.

\paragraph{Refinement.}
Finally, we fit a Gaussian-splat scene~\cite{kerbl2023gaussian} to the video, with Gaussians anchored
to the object and hand meshes (and to a background point cloud), and the camera is initialized
from monocular SLAM (DROID-SLAM~\cite{teed2021droid}). Optimizing photometric, segmentation,
and depth losses under temporal-smoothness penalties jointly refine and smooth the
object pose and scale, hand pose and scale, and camera poses that constitute the final
reconstruction.

\begin{figure}[h]
    \centering
    \begin{minipage}[b]{0.32\linewidth}
        \centering
        \textbf{(a) Unaligned}\\[4pt]
        \includegraphics[width=\linewidth]{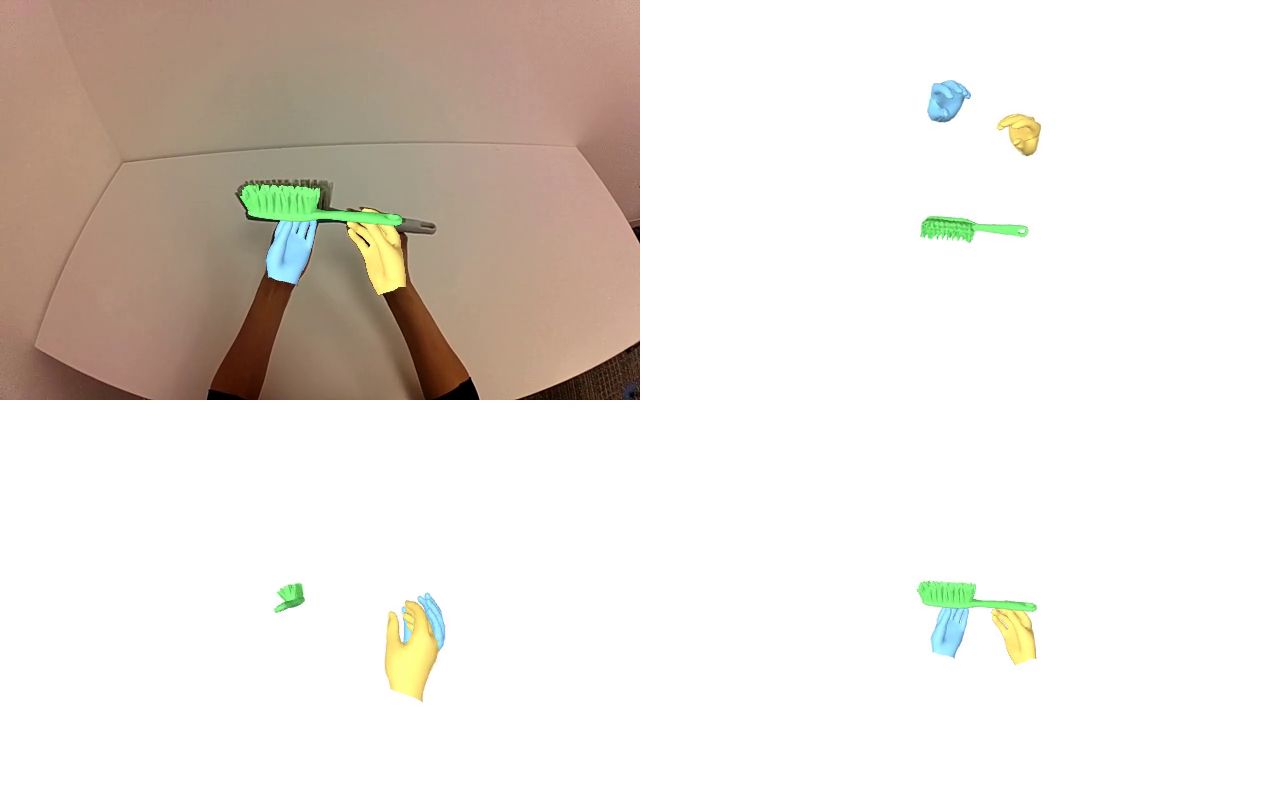}
    \end{minipage}
    \hfill
    \begin{minipage}[b]{0.32\linewidth}
        \centering
        \textbf{(b) Aligned}\\[4pt]
        \includegraphics[width=\linewidth]{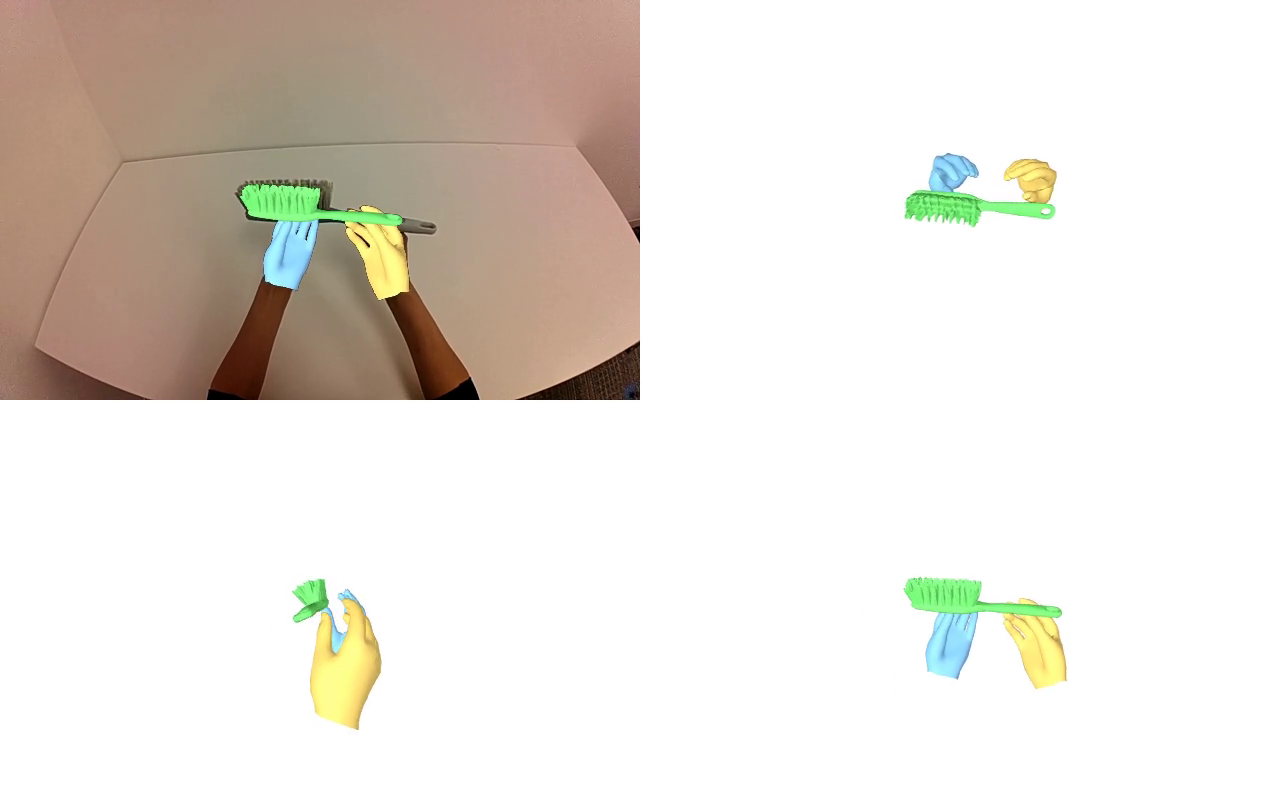}
    \end{minipage}
    \hfill
    \begin{minipage}[b]{0.32\linewidth}
        \centering
        \textbf{(c) Refined}\\[4pt]
        \includegraphics[width=\linewidth]{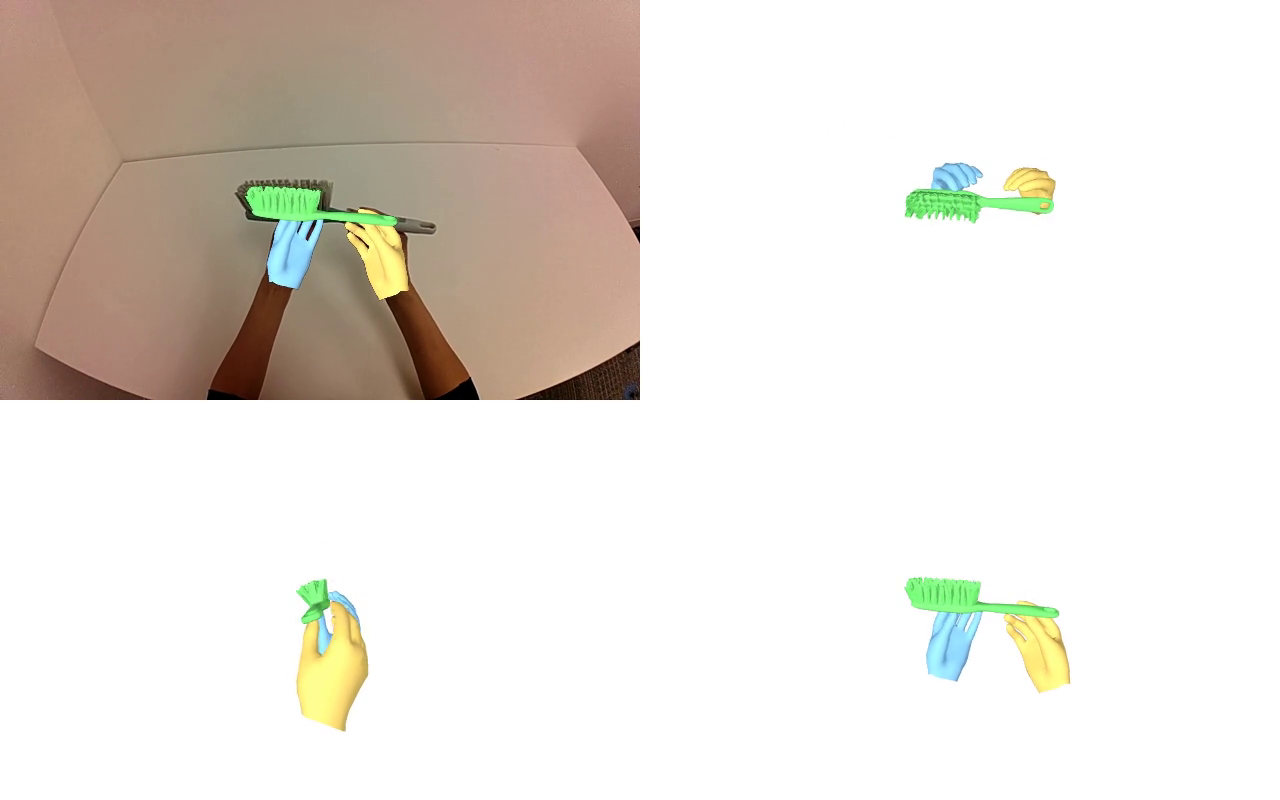}
    \end{minipage}
    \caption{Hand--object reconstruction across three stages of processing.
    Each panel shows the source camera view (top-left) and three world-frame views.
    \textbf{(a)} Unaligned: raw Dyn-HaMR output before any correction.
    \textbf{(b)} Aligned: after depth alignment with MoGe monocular depth estimates.
    \textbf{(c)} Refined: after Gaussian splat-based joint optimization.}
    \label{fig:moge_renders}
\end{figure}

\begin{figure}[h]
    \centering
    \includegraphics[width=\linewidth]{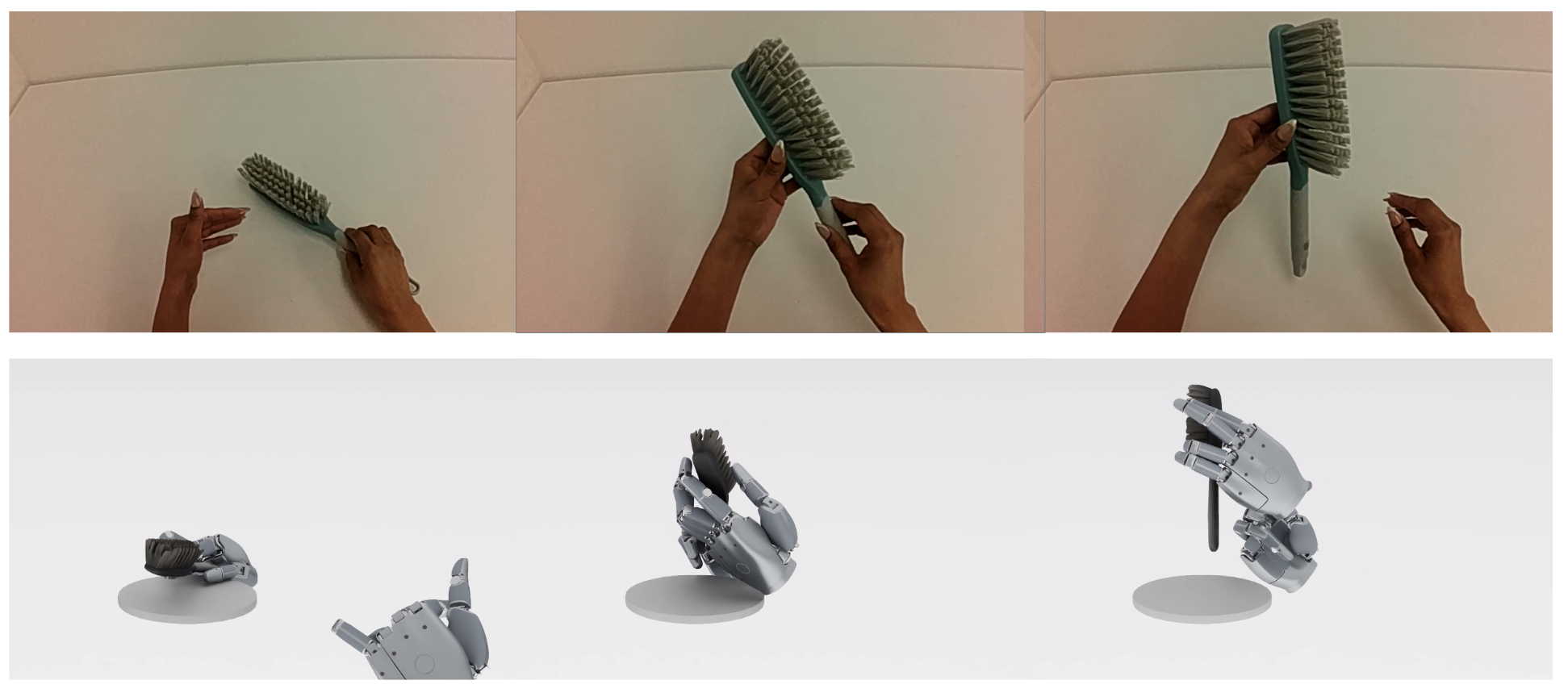}
    \caption{Hand--object reconstruction result after the full reconstruction pipeline.}
    \label{fig:end_to_end_validation}
\end{figure}

Figure~\ref{fig:end_to_end_validation} visualizes the demonstration video and resulting grounded behavior. The top row shows representative frames from the original human demonstration, while the bottom row shows the corresponding grounded robot execution in simulation. Despite substantial embodiment differences between the human demonstrator and robot, \algabrvname{} successfully transfers the demonstrated interaction and produces physically plausible manipulation behavior.

To quantitatively evaluate grounding quality, we report $0.647$ AUC demonstrating that the proposed pipeline can transform noisy human video observations into executable robot trajectories. These results validate the complete system, including human motion reconstruction, contact extraction, contact-aware optimization, and embodiment grounding, highlighting the practicality of \algabrvname{} for real-world human-to-robot behavior transfer.

\section{Dexterous Manipulation Benchmark Details}
\label{app:task_library}

\paragraph{Post-process.}
Our task library converts human hand-object interaction data into Sharpa-Wave robot trajectories. It ingests per-frame MANO hand fits and rigid-object 6-DoF trajectories from seven motion-capture datasets: ARCTIC, TACO, HOT3D, OakInk2, DexYCB, GRAB, and H2O. Each dataset-specific loader parses the source’s native format and exports every sequence into a unified representation containing the wrist pose, 21 MANO joint positions, per-link hand-object contacts, and the object’s full 6-DoF pose.

For HOT3D, the long egocentric recordings are first segmented into atomic hand-object interaction clips before retargeting. Contact is detected when the minimum fingertip-to-object distance falls below 2 cm. Brief contact breaks of up to 15 frames are bridged, and each resulting contact window is split according to the dominant contacted object using a per-frame majority vote. This process expands roughly 294 raw recordings into approximately 4,045 atomic clips.

Each sequence is then retargeted to the robot using differential inverse kinematics solved with a quadratic-programming optimizer. The per-frame objective tracks the human wrist position and orientation, as well as fingertip residuals, while respecting the robot’s joint-position and joint-velocity limits. Optimization runs for up to 200 iterations per frame and terminates early when the residual change falls below 1e-6.

\paragraph{Quality Check.}
After retargeting, every sequence undergoes two complementary quality checks to determine whether it is accepted into the training set or rejected. The penetration check models the robot hand as a set of capsules and measures both hand-object overlap and hand-hand overlap on each frame. Hand-object overlap is computed against the convex hull of the object, while hand-hand overlap uses analytic segment-to-segment distances between capsule pairs. A sequence is rejected if either penetration measure exceeds 2 cm in any frame. Highly concave objects, defined as having a convex-hull-to-mesh volume ratio above 3, are excluded from the hand-object penetration check because their convex hulls can falsely classify the hand as being inside open objects such as mugs or AR glasses.

The replay check validates each sequence in the Isaac Lab simulator by replaying the trajectory with tracking actions for 300 steps. This verifies that the scene initializes correctly and that the simulator can execute the trajectory without failure. A sequence is marked successful only after the rendered video is fully written to disk. As a result, any simulator issue, including CUDA asserts, NaN observations, physics instabilities, joint-limit violations, Omniverse shutdown deadlocks, or video-encoding failures, prevents the success marker from being recorded.

\paragraph{Dataset Statistics.} To characterize the contact behavior in our task library with human motions, we report the following metrics, computed from the precomputed contact wrench matrix $\mathcal{W}_{h,k}$ and its support values $h_{h,k}$ along $512$ basis directions.
  
\begin{itemize}[leftmargin=*, itemsep=2pt, topsep=2pt]
% \item \textbf{Contact-frame ratio}: fraction of frames in which at least one hand is in active contact. Indicates how much of the motion is true manipulation versus transit.
\item \textbf{Contact events}:  number of contact intervals per hand. Together with the \emph{segment duration} distribution, it distinguishes long contact events from frequent short touches.
% \item \textbf{Bimanual ratio}:  fraction of frames where both hands are simultaneously in contact, reported both unconditionally and restricted to the \emph{same object body} (cooperative dual-arm manipulation).
% \item \textbf{Wrench-space spread} $\mathrm{std} (h_{h,k})$:  anisotropy of the achievable wrench polytope; higher values denote stronger directional preference and the complexity of the wrench space manifold.
\item \textbf{Ferrari--Canny epsilon} $\varepsilon=\min h_{h,k}$:  distance from the origin to the nearest face of the convex wrench hull in wrench matrix; $\varepsilon > 0$ certifies force closure and measures disturbance resistance.
% \item \textbf{Force--torque split} $\|u_{1,1:3}\|^2$:  fraction of the unit principal axis $u_1$ associated with $s_1$ (from per-frame uncentered SVD) carried by the force block; indicates whether dominant wrenches are translational or rotational.
% \item \textbf{PC1 variance fraction} $s_1 / \sum_k s_k$:  concentration of the wrench rays along a single 6D direction; close to $1$ for near rank-1 contacts with more dynamics contact behavior, close to $1/6$ for isotropic full-hand grasps.
% \item \textbf{Principal-axis diversity}:  mean pairwise cosine distance between $|u_1|$ vectors across all $(\text{frame}, \text{hand}, \text{body})$ tuples within a sequence; low values indicate stereotyped single-mode interactions, high values indicate rich contact behaviors.
\end{itemize}

\section{Extending CHORD to Whole Body Manipulation}
\label{app:whole_body}

We propose a generic pipeline for extending CHORD to whole-body manipulation dependent on the human demonstration data available. Our pipeline leverages a complete or synthetically generated whole body reference trajectory and learns a residual following the CHORD recipe to perform whole body manipulation. We discuss our approach for generating a whole body reference trajectory given a hand-only reference (ex. obtained from egocentric video) in Sec. \ref{subsec:hand-only-ref}. We then discuss how we apply CHORD using a whole body reference trajectory in Sec. \ref{subsec:whole-body-chord}.

\subsection{Planning Whole Body Motion from End Effector Trajectories}
\label{subsec:hand-only-ref}

\begin{figure}
    \centering
    \includegraphics[width=\linewidth]{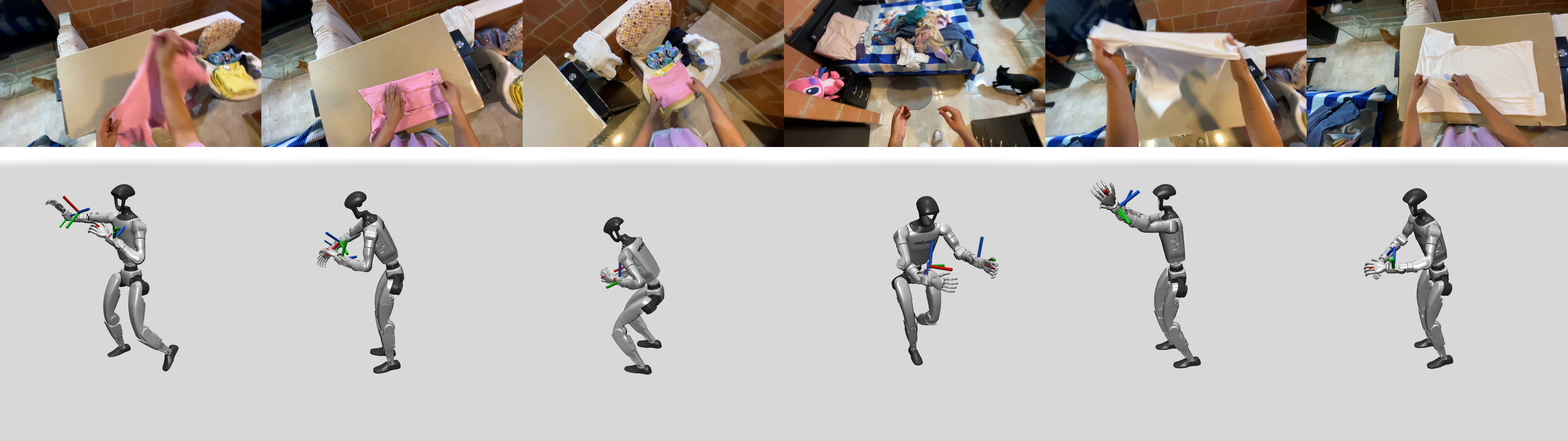}
    \caption{Planned whole body motion for end effector trajectories extracted from egocentric video.}
    \label{fig:placeholder}
\end{figure}

To leverage hand-only references for learning whole body manipulation, we first retarget the reconstructed MANO hand to the target robotic dexterous hand. Given the retargeted end effector trajectories, we seek a way to generate a corresponding whole body motion that can be tracked by a whole body controller \cite{luo2025sonic}. To this end, we train a end-effector to whole-body inpainting module, utilizing the architecture proposed in \cite{wang2026motionbricksscalablerealtimemotions}, largely reusing their proposed notation for clairty.

\paragraph{Constraint Specification.} The keyframe constraints $\mathcal{T} = \{\mathcal{T}_1, \mathcal{T}_2, \mathcal{T}_3\}$ provide sparse, optional anchors on the local root, global root, and pose respectively. Our specialisation fixes $\mathcal{T}_1, \mathcal{T}_2$ to be empty (i.e. masked at every temporal slot) and restricts $\mathcal{T}_3$ to the wrist body subset, populated at \emph{every} frame. Let, $\mathcal{H} \subset \mathbb{B}$ index the EE body slots ($\mathcal{H} = \{\mathrm{L},\mathrm{R}\}$ for the left and right wrist), and let
\begin{equation}
\mathcal{T}^{\text{EE}} \;=\; \bigl\{\,\bigl(p_h^{\star,t},\,q_h^{\star,t}\bigr) \,\bigm|\, t \in [1,T],\ h \in \mathcal{H}\,\bigr\}
\label{eq:ee_constraint_set}
\end{equation}
be the dense EE constraint set, where $p_h^{\star,t} \in \mathbb{R}^3$ is the target wrist position in the global root-relative frame and $q_h^{\star,t}$ is its rotation in 6D.

\paragraph{Training Data and Constraint Sampling.} We train the tokenizer, root module, and pose module on the retargeted BONES Seed dataset. Motion segments are sampled following \cite{wang2026motionbricksscalablerealtimemotions}. For each segment we extract the ground-truth EE conditioning,
\begin{equation}
\mathcal{T}^{\text{EE}}_{\text{gt}} \;=\; \bigl\{\,\bigl(p_h^{t},\,q_h^{t}\bigr) \,\bigm|\, t \in [1,T],\ h \in \mathcal{H}\,\bigr\}
\label{eq:ee_gt_extract}
\end{equation}
and with probability $\rho_{\text{drop}}$ each $(h, t)$ entry is replaced by a learnable mask embedding.

\paragraph{Tokenizer Training.} The tokenizer is trained as described in \cite{wang2026motionbricksscalablerealtimemotions} with the addition of an auxiliary end effector loss.

\paragraph{Root Module Training.} Given the keyframe constraints $\mathcal{T}^{\text{EE}}_{\text{gt}}$, the root module is trained to predict:
\begin{enumerate}[leftmargin=1.5em,noitemsep,topsep=0pt]
    \item ($\mathcal{F}_1$): predict the in-betweening frame count $T_2$ as described in \cite{wang2026motionbricksscalablerealtimemotions}, optimized with cross entropy loss on binned frame counts.
    \item ($\mathcal{F}_2$): predict the global root trajectory conditioned on $T_2$, as described in \cite{wang2026motionbricksscalablerealtimemotions}, optimized with smooth-$\ell_1$loss on the ground truth root trajectory.
\end{enumerate}

\paragraph{Pose Module Training.} Given the keyframe constraints $\mathcal{T}^{\text{EE}}_{\text{gt}}$, we transform them to be root relative given the ground truth root trajectory and then train the pose module to predict the discrete tokenizer codebook indices corresponding to the ground truth whole body motion, optimized with cross entropy loss.

\paragraph{Inference Loop.} At inference, the planner consumes (i)~a per-frame EE target trajectory $\mathcal{T}^{\text{EE}}$ of total length $T_{\text{seq}}$ and (ii)~$T_{\text{ctx}} = 4$ context frames seeding the autoregressive rollout. It produces the full-body state sequence $\{r_g^t, r_l^t, p^t, q^t, v^t, c^t\}_{t=1}^{T_{\text{seq}}}$. Generation proceeds chunk by chunk in MotionBricks tokens. The autoregressive seed for chunk $j$ is provided by the last $T_{\text{ctx}}$ frames of chunk $j{-}1$. For chunk $0$ the seed is supplied as a nominal pose is not otherwise available. In practice, we inference the planner to produce overlapping chunks and apply blending to the decoded motions.

\subsection{Applying CHORD}
\label{subsec:whole-body-chord}
Given a whole body reference we learn a residual on top of a pretrained whole body controller \cite{luo2025sonic} to adjust the controller's actions to induce the desired object trajectory by aligning the robot's contact wrench to the human demonstrated contact wrench. Our implementation details largely follow Sec. \ref{app:rl} with notable differences as follows,

\paragraph{Observation Space.} We make the following additions to the observation space,
\begin{itemize}[leftmargin=1.5em,noitemsep,topsep=0pt]
    \item \textbf{Joint state}: whole body joint positions and velocities.
    \item \textbf{Command}:
    \begin{itemize}[leftmargin=1.5em,noitemsep,topsep=0pt]
        \item \textbf{Anchor}: target deltas for the position and orientation of the anchor body from the current frame to N state reference frames.
        \item \textbf{End Effectors}: target deltas for the position and orientation of the end effector bodies from the current state to N future reference frames.
        \item \textbf{Joint Position}: target deltas for the joint position from the current state to N future reference frames.
    \end{itemize}
\end{itemize}
We remove \textbf{contact observation} from the whole body training observation space.

\paragraph{Action Space.} The policy controls residual actions for the whole body joint positions. The residual actions are applied to the output of a whole body controller tracking the original whole body reference with an action scale of 0.5 for non-finger joints and 0.15 for finger joints. These adapted actions are then used as joint position targets.

\paragraph{Termination and Reset.} We add the following terminations:
\begin{itemize}[leftmargin=1.5em,noitemsep,topsep=0pt]
    \item \textbf{Anchor Position Error}: the anchor deviates too far from the commanded anchor position.
    \item \textbf{Anchor Orientation Error}: the anchor deviates too far from the commanded anchor orientation.
\end{itemize}

To reset at arbitrary points the trajectory, we leverage the stability of the whole body controller where the environment restores the robot state and object state at a randomly sampled frame in the trajectory and holds the commanded reference at this frame for N frames. During this period, selected joints are "opened" using a offset factor applied to the nominal joint positions in the reference frame and the per-environment virtual object control is held at scale 1.0 for M < N frames before decaying to the training curriculum scale. In our experiments, we use an offset factor of 0.5 on the shoulder yaw joints of the G1 to pull the arms back from the manipulated object(s) and allow the policy to learn to recover towards the object(s).

\paragraph{Additional Rewards} We add the following rewards,
\begin{itemize}[leftmargin=1.5em,noitemsep,topsep=0pt]
    \item \textbf{Anchor Position Reward:} rewards tracking the reference anchor position with a gaussian error kernel.
    \item \textbf{Anchor Orientation Reward:} rewards tracking the reference anchor orientation with a gaussian error kernel.
\end{itemize}

\subsection{Contact Reward}
We ablate the role of our proposed contact wrench reward in learning whole body manipulation by comparing to an identical training recipe using contact position rewards (as used in DexMachina). As our reference contact is from a five finger human-hand and our target embodiment is a G1 humanoid with the Dex3 three fingered hand, this ablation further questions the ability of our contact wrench reward to support cross-embodiment transfer to different hand morphologies. As seen in Table \ref{tab:ablate-whole-body-contact-rew}, while contact position rewards suffer when transferring to a different embodiment, wrench space alignment guides learning towards contact behaviors that reproduce the induced object motion. We reason this is due to contact position rewards being tightly coupled with the demonstrated embodiment, whereas wrench space alignment guides learning independently of embodiment constraints.

\begin{table}[t]
  \centering
  \begin{tabular}{llcc}
    \toprule
    Category & $n$ & Wrench Reward (SR) & Position Reward (SR)  \\
    \midrule
    Rigid        &  4 & \textbf{$0.994 \pm 0.008$} & $0.460 \pm 0.487$ \\
    Articulated  &  4 & \textbf{$0.914 \pm 0.107$} & $0.000 \pm 0.000$ \\
    Multi-object &  4 & \textbf{$0.866 \pm 0.129$} & $0.192 \pm 0.166$ \\
    \midrule
    Overall         & 12 & \textbf{$0.925 \pm 0.104$} & $0.217 \pm 0.333$ \\
    \bottomrule
    \end{tabular}
    \caption{Performance comparison across different reward formulations. Values are averaged across 4096 evaluation episodes and 4 seeds per sequence for each of the $n$ sequences in each category. Best values are bolded.}
    \label{tab:ablate-whole-body-contact-rew}
\end{table}

\section{Hardware Setup}
\label{app:hardware}

\paragraph{Hardware Platform.}
Our real-world experiments are conducted on a Dexmate robot equipped with two Sharpa dexterous hands. The deployment controller actuates the left and right Dexmate arms, the torso, and both Sharpa hands. Each arm has seven actuated joints, the torso has three actuated joints, and each Sharpa hand provides 22 actuated joints, giving a 61-dimensional commanded joint vector during deployment. The robot head and mobile base are not used by the policy controller in these experiments. The Dexmate torso is initialized and held at a fixed crouched configuration, $q_{\mathrm{torso}}=[0.78, 1.57, 0.44]rad$. The low-level bridge publishes robot state at 100 Hz and sends joint commands with a zero-order hold at up to 500 Hz.
Object and robot poses are measured with a six-camera Vicon motion-capture system. Adhesive reflective markers are attached to the robot and manipulated objects to define trackable rigid bodies.

\paragraph{Vicon Setup and Calibration Pipeline}
\label{app:vicon}
The Vicon system streams rigid-body poses to the deployment process through a ZMQ publisher with 100 Hz. Each Vicon frame contains a timestamped set of subjects, where each subject pose is represented by a 3D position and a unit quaternion in $wxyz$ convention. The deployment process subscribes to this stream, keeps the latest frame, and rejects stale frames older than 0.25$s$.
We use the Vicon-tracked robot segment to align the Vicon world frame with the robot kinematic model. The tracked robot segment is associated with the \texttt{torso\_l3} frame in the Dexmate URDF. Given the current robot joint state, forward kinematics computes the transform from the robot base to this tracked frame. Combining this kinematic transform with the Vicon measurement yields the robot base pose in the Vicon world frame.
Before policy execution, we calibrate the frame alignment using the object pose. We denote the Vicon world frame by $V$, the robot base frame by $B$, the policy environment frame by $E$, and the object frame by $O$. The policy predicts wrist targets in $E$, while the real robot controller and inverse kinematics operate in $B$. Therefore, we first solve the environment-to-base transform $T^B_E$. Let $T^B_O$ be the live object pose expressed in the robot base frame, and let $T^E_O$ be the corresponding object pose in the recorded motion-reference trajectory used by the policy. Since both poses describe the same physical object configuration at initialization, they satisfy $T^B_O = T^B_E T^E_O$, giving
\begin{equation}
    T^B_E = T^B_O (T^E_O)^{-1},
\end{equation}
After this one-time startup calibration, $T^B_E$ is held fixed and used to map policy wrist targets from the policy frame into the robot base frame before inverse kinematics.

The policy observation, however, also needs live wrist and object poses expressed in $E$, while Vicon measures those poses in $V$. To avoid routing every live Vicon pose through the robot base frame, we compute a second fixed transform $T^E_V$ at the same calibration instant. Let $T^V_O$ be the live object pose measured directly by Vicon. Because $T^E_O$ and $T^V_O$ correspond to the same initialized object pose, we set
\begin{equation}
    T^E_V = T^E_O (T^V_O)^{-1}.
\end{equation}
At runtime, any live Vicon pose $T^V_X$ is converted to the policy frame as $T^E_X = T^E_V T^V_X$, where $X$ may be the object or either wrist frame.

\paragraph{Policy Deployment Workflow}
The deployed policy is exported as an ONNX model and executed with ONNX Runtime inside a real-time inference loop. The loop runs at the trajectory rate, 20 Hz in our experiments, while the lower-level hardware bridges hold the most recent command at a higher frequency.
% Each policy step follows the same sequence. First, the deployment process reads the current robot state and converts it into a simulation-frame observation using forward kinematics and the calibrated frame transform. If Vicon is enabled, the live wrist and object poses replace the purely kinematic/object trajectory estimates in the policy observation. The ONNX policy then predicts a 56-dimensional action consisting of right- and left-hand commands. For each hand, the action contains a 3D wrist position delta, a 3D wrist orientation delta, and 22 finger joint deltas.
% The action is post-processed to match the training-time controller. Raw deltas are scaled, clipped, filtered with an exponential moving average, and accumulated into absolute wrist and finger targets. Finger targets are clamped to the Sharpa joint limits. The resulting wrist targets are transformed into the robot base frame, solved through inverse kinematics, and sent to the hardware as joint commands.
The inference loop uses a finite-state machine with \textsc{Hold}, \textsc{Play}, \textsc{Safety}, and \textsc{Done} states. In \textsc{Hold}, the robot maintains the calibrated initial pose. Pressing the operator start key resets the policy action accumulator from the current robot state and enters \textsc{Play}. If Vicon tracking becomes stale or the robot marker body violates the drift threshold, the system enters \textsc{Safety}, freezes the current joint state, and requires an operator reset before continuing.

\section{Comparing \algabrvname to Teleoperation}
\label{app:Teleop_VS_RL}

We conducted a small qualitative teleoperation pilot study to better understand the difficulty of reproducing our demonstration trajectories with a human operator in the loop. In each trial, the operator first watched a demonstration video of the target task and then attempted to reproduce the same behavior through the teleoperation interface. We tested several contact-rich manipulation tasks, including the box task, mixer task, and waffle-iron task. 

The study revealed several consistent limitations of the teleoperation setup for contact-rich dexterous manipulation. The box task was especially difficult, as successful execution required precise coordination between grasp formation, contact timing, and force application. Across tasks, we observed three primary sources of difficulty. First, the hand retargeting module did not always preserve the operator’s intended finger configuration, particularly for wide hand apertures or poses near the robot hand’s kinematic limits. Second, because the operator received no haptic or contact feedback, the contact state had to be inferred solely from visual observation, which was often insufficient for accurately regulating interaction with the object due to the occlusions. Third, the teleoperation has dufficulty of offering direct control over joint-level torques, limiting the operator’s ability to deliberately apply force through a particular finger or joint. 

For the mixer and waffle-iron tasks, operators could often identify a strategy that produced an approximately correct object trajectory after extended practice. However, these executions still differed from the demonstrations in the underlying contact interactions. In particular, the resulting grasps were fragile and sensitive to small errors in contact location, timing, or applied force, which frequently led to unstable manipulation or task failure.

\section{Baseline Reproduction Implementation Details}
\label{app:baselineImplementation}

To produce the results reported in \Cref{tab:comparison_prior_work}, we extended the prior methods DexMachina, ManipTrans, and Spider with Sharpa hands and leveraged their original code base. In this section, we describe our implementation efforts to faithfully reproduce these methods on the new embodiment. Notably, both DexMachina and ManipTrans claim to be embodiment-agnostic and capable of supporting arbitrary robot embodiments.

\subsection{DexMachina}
We first follow the dexterous hand asset processing instructions provided by the authors of DexMachina to add the Sharpa Hand. This includes running the provided urdf inspection script, adding 6 DOF joints to the wrists, and creating the retargeting configs mapping the robot fingers to MANO keypoints. We then specify collision groups using the provided utilities and add a corresponding configuration python file for the Sharpa Hand. We tune the controller gains following the procedure described in the documentation and also keep an alternate configuration copying the gains used in our IsaacLab CHORD training. Finally, we run retargeting on the provided sequences as well as our own and visually inspect each retargeting in simulation kinematic playback to verify the hand trajectory, object trajectory, and contact markers are all consistent with the MANO demonstration.

After asset verification and retargeting, we use the released DexMachina RL training recipe as implemented but with the Sharpa Hand asset and previously retargeted sequences. We run 4 variants of training with 3 seeds each for all sequences. We train (1) using the controller gains tuned in DexMachina with Hybrid Action space (2) using the controller gains used in CHORD IsaacLab with Hybrid Action Space (3) using the controller gains tuned in DexMachina with Residual Action Space (4) using the controller gains in CHORD IsaacLab with Residual Action Space. We report the best performing variant (1).

\subsection{Maniptrans}
We reimplemented ManipTrans's manipulation curriculum in our training
framework, following the baseline adaptation approach described in
DexMachina. We trained per-sequence policies for the Sharpa hands without
the original pretrained hand-motion imitator and its separate residual
module. The ManipTrans results in \Cref{tab:comparison_prior_work}
correspond to this adapted implementation.

\subsection{Spider}
We first reproduced the experiments from the original Spider codebase to verify that our results were consistent with those reported in the paper. We then integrated the Sharpa Hands model into the baseline pipeline. To generate robot motions, we used the \algabrvname IK solver and also evaluated several alternative IK solutions with extensive parameter tuning, but none produced consistently satisfactory results. We manually inspected the generated IK motions to ensure that they were visually plausible and functionally feasible for the target tasks. Finally, we ran Spider on the selected tasks using 3 random seeds per task and recorded the resulting errors for analysis.

\subsection{Human2Sim2Robot}
We reimplemented Human2Sim2Robot's training recipe in our framework, following the adaptation procedure described in DexMachina.

\end{document}